\documentclass[final]{cvpr}

\usepackage{times}
\usepackage{epsfig}
\usepackage{graphicx}
\usepackage{amsmath}
\usepackage{amssymb}
\usepackage{float}

\usepackage[utf8]{inputenc}
\usepackage{url}
\usepackage{subcaption}
\usepackage{booktabs}
\usepackage{wasysym}
\usepackage{pifont}
\usepackage{xspace}
\usepackage{multirow}
\usepackage{microtype}

\usepackage{amssymb}
\usepackage{amsmath,amsfonts}
\usepackage{amsopn,bm,mathtools}

\usepackage{nicefrac}       

\newcommand{\ProbOpr}[1]{\mathbb{#1}}

\newcommand{\expect}[2]{%
\ifthenelse{\equal{#2}{}}{\ProbOpr{E}_{#1}}
{\ifthenelse{\equal{#1}{}}{\ProbOpr{E}\left[#2\right]}{\ProbOpr{E}_{#1}\left[#2\right]}}} 
\newcommand{\var}[2]{%
\ifthenelse{\equal{#2}{}}{\ProbOpr{VAR}_{#1}}
{\ifthenelse{\equal{#1}{}}{\ProbOpr{VAR}\left[#2\right]}{\ProbOpr{VAR}_{#1}\left[#2\right]}}} 

\usepackage{xspace}

\makeatletter
\DeclareRobustCommand\onedot{\futurelet\@let@token\@onedot}
\def\@onedot{\ifx\@let@token.\else.\null\fi\xspace}
\def\eg{\emph{e.g}\onedot} 
\def\ie{\emph{i.e}\onedot} 
\def\cf{\emph{c.f}\onedot}

\makeatother

\newcommand\brackets[1]{\left[#1\right]}

\usepackage{mathtools}

\usepackage[ruled,vlined]{algorithm2e}

\usepackage{listings}

\graphicspath{{../figs/}{../}}

\newcommand{\eat}[1]{}

\providecommand{\ourtitle}{Embedding Adaptation is Still Needed for Few-Shot Learning}
\providecommand{\ourmethod}{\textsc{ATG}\xspace}
\providecommand{\codelink}{http://seba1511.net/projects/atg}

\newcommand{\x}{\mathbf{x}}
\DeclareMathOperator*{\E}{\mathop{\mathbb{E}}}
\newcommand{\Exp}[2][]{\E_{#1}\brackets{#2}}
\newcommand{\softmax}{\text{softmax}}
\newcommand{\h}[1]{\textbf{#1}}
\newcommand{\e}[1]{#1}
\newcommand{\ptrain}{p_\text{train}}
\newcommand{\ptest}{p_\text{test}}
\newcommand{\mutrain}{\mu_\text{train}}
\newcommand{\mutest}{\mu_\text{test}}

\makeatletter
\@namedef{ver@everyshi.sty}{}
\makeatother
\usepackage[colorinlistoftodos,linecolor=gray,backgroundcolor=none,bordercolor=red,textsize=small,textwidth=16mm]{todonotes}
\definecolor{fscolor}{rgb}{0, 0, 1}

\usepackage[pagebackref=true,breaklinks=true,colorlinks,bookmarks=false]{hyperref}

\usepackage[pagebackref=true,breaklinks=true,colorlinks,bookmarks=false]{hyperref}

\def\cvprPaperID{6231} 
\def\confYear{CVPR 2021}

\begin{document}

\title{\ourtitle}

\author{
S\'ebastien M. R. Arnold\\
University of Southern California\\
{\tt seb.arnold@usc.edu}
\and
Fei Sha\\
Google AI\\
{\tt fsha@google.com}
}

\maketitle

\begin{abstract}
Constructing new and more challenging tasksets is a fruitful methodology to analyse and understand few-shot classification methods.
Unfortunately, existing approaches to building those tasksets are somewhat unsatisfactory: they either assume train and test task distributions to be identical --- which leads to overly optimistic evaluations --- or take a ``worst-case'' philosophy --- which typically requires additional human labor such as obtaining semantic class relationships.
We propose \ourmethod, a principled clustering method to defining train and test tasksets without additional human knowledge.
\ourmethod models train and test task distributions while requiring them to share a predefined amount of information.
We empirically demonstrate the effectiveness of \ourmethod in generating tasksets that are easier, in-between, or harder than existing benchmarks, including those that rely on semantic information.
Finally, we leverage our generated tasksets to shed a new light on few-shot classification: gradient-based methods --- previously believed to underperform --- can outperform metric-based ones when transfer is most challenging.
\end{abstract}


\section{Introduction}
\label{sIntroduction}

Few-shot learning, the ability to learn from limited supervision, is essential for the real-world deployment of adaptive machines.
Although proposed more than 20 years ago~\cite{Miller2000-hy, Fei-Fei2006-od}, this field has recently been the focus of vast research efforts and a plethora of methods were proposed to tackle many of its challenges, including knowledge transfer and adaptation.

However, the fundamental problem of \emph{evaluating} those methods remains largely unaddressed.
Although many standardized benchmarks exist, they follow one of two recipes to generate classification tasks -- they either partition classes at random (\eg~\cite{Bertinetto2018-gn, Vinyals2016-cm}) or leverage class semantic relationships (\eg~\cite{Oreshkin2018-yi, Ren2018-hk, Triantafillou2019-bj}).
The former implicitly assumes that train and test tasks come from the same distribution, leading to overly optimistic evaluation.
The latter, although more realistic, requires additional human knowledge which can be expensive to gather, when available at all.
This status quo is unsatisfactory because different applications call for different benchmarking schemes: a model that performs best when train and test tasks are similar does not necessarily achieve top accuracy when the two tasksets significantly differ.
In other words, the quality of a few-shot learning algorithm depends on both train and test tasks, and \emph{their relative similarity}.


\begin{table}[t]
    \vspace{0em}
    \centering
    {\small
    \centering
    \caption{
        \small
        5-ways 5-shots classification accuracy of metric- and gradient-based methods when transfer is most challenging.
        In this regime, methods that adapt their embedding function (Finetune, MAML) outperforms those that do not, and which were thought to be sufficient for few-shot learning.
    }
    \label{tHardPartitions}
    \vspace{-1em}
    \setlength{\tabcolsep}{3pt}
    \begin{tabular}{@{}l ccccc@{}}
        \toprule

                        & CIFAR100    & mini-IN     & tiered-IN   & LFW10       & EMNIST \\
        \midrule
        ANIL            & 49.27\%     & 54.42\%     & 73.24\%     & 77.86\%     & 86.54\%  \\
        ProtoNet        & 50.46\%     & 60.36\%     & 78.64\%     & 85.48\%     & 89.81\%  \\
        Multiclass      & 53.02\%     & 61.86\%     & 81.90\%     & 82.26\%     & 90.96\%  \\
        MAML            & 55.73\%     & 62.25\%     & n/a         & \h{86.36\%} & 91.92\% \\
        Finetune        & \h{70.98\%} & \h{70.47\%} & \h{83.57}\% & 83.35\%     & \h{93.51\%}  \\

        \bottomrule
    \end{tabular}
    }
    \vspace{-1em}
\end{table}

Without more fine-grained benchmarks, we might miss important properties of our algorithms and hinder their deployment to real-world scenarios.
For example, recent work suggests that simply learning a good feature extractor might be \emph{all we need} for few-shot classification~\cite{Tian2020-gv, Raghu2019-ff}.
Notably, they match and often surpass the performance of gradient-based methods by sharing a feature extractor across all tasks and only adapting a final classification layer.
But, it stands to reason that in the extreme case where train tasks contain no information relevant to the test ones (\ie transfer is impossible), those methods will underperform those that are allowed to adapt their feature extractor.
Table~\ref{tHardPartitions} diplays results for such an instances where gradient-based methods (MAML~\cite{Finn2017-gw}) dominate on tasksets carefully designed to challenge transfer.


\begin{figure*}[ht!]
    \vspace{-2em}
    \begin{center}
        \includegraphics[width=0.9\linewidth]{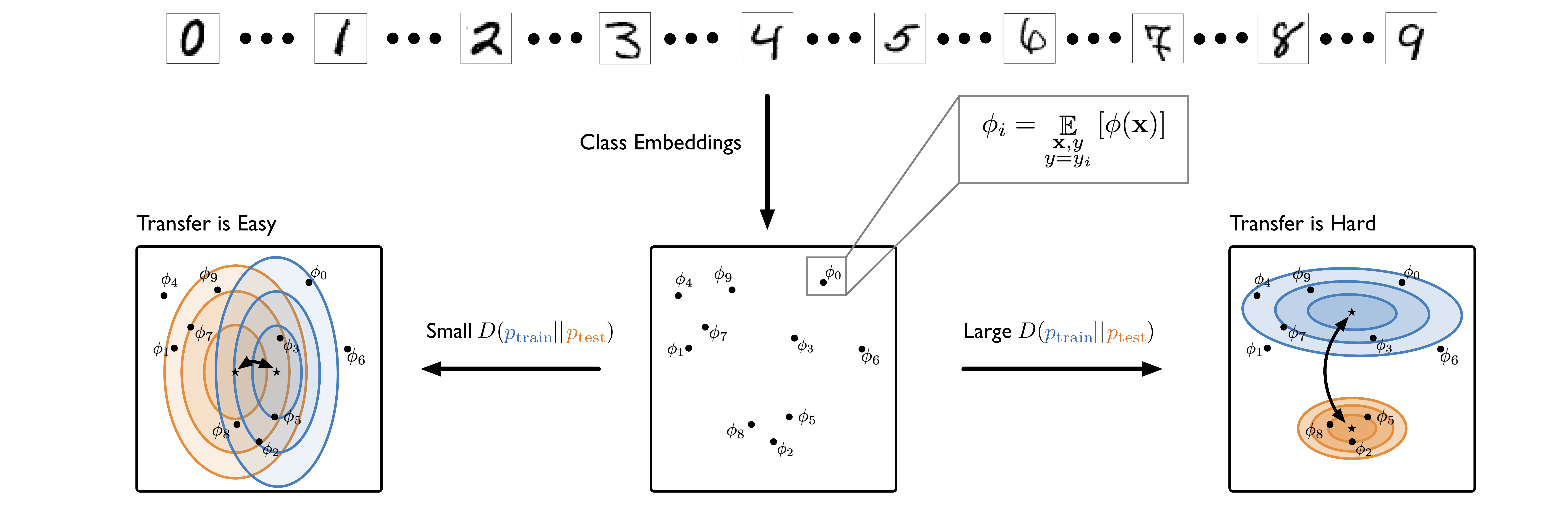}
    \end{center}
    \vspace{-1em}
    \caption{
        \small
        \ourmethod, a method to generate tasksets of varying difficulty.
        First, we compute each \emph{class embedding} by averaging the embedding $\phi(\x)$ of all images $\x$ associated with that class.
        Then, we partition those class embeddings using a penalized clustering objective.
        If we want easy tasksets, we find clusters such that train and test classes are pulled together; for hard tasksets, we push those distributions apart.
    }
    \label{fPipeline}
    \vspace{-1em}
\end{figure*}

In this manuscript we propose \emph{Automatic Taskset Generation} (\ourmethod), a method to automatically generate tasksets from existing datasets, with fine-grained control over the transfer difficulty between train and test tasks.
(\cf Figure~\ref{fPipeline})
Our method can be understood as a penalized clustering objective that enforces a desired similarity between train and test tasks.
Importantly, it does not require additional human knowledge and is thus amenable to settings where this information is not available.
We use \ourmethod to study and evaluate the two main families of few-shot classification algorithms: gradient-based and metric-based methods. 
Our results on 5 tasksets, including two new ones, show that gradient-based method become particularly compelling when transfer is most challenging.

\paragraph{Contributions}
We make the following contributions:
\begin{itemize}
    \item \ourmethod, a method to automatically generate tasksets that does not require additional human knowledge.
    \item Extensive validation and study of our method, showing it can effectively control the degree of transfer difficulty between train and testk tasks.
    \item An empirical analysis of popular few-shot learning methods, suggesting that gradient-based method outperform metric-based ones in the most challenging transfer regimes.
\end{itemize}
Our implementations and tasksets are described in the Supplementary Material, and are publicly available at:

\noindent
\url{\codelink}


\section{Related Works}
\label{sRelatedWorks}

\paragraph{Few-Shot Learning}
The goal of few-shot learning is to produce a model able to solve new tasks with access to only limited amounts of data~\cite{Fei-Fei2006-od, Miller2000-hy}.
It is closely related to meta-learning -- devising models that learn to learn~\cite{Bengio1991-yp, Schmidhuber1987-cx} -- but with a particular emphasis on the small quantity of available data.
This research direction has received a lot of interest in recent years, due to the numerous applications in natural language processing~\cite{Mu2019-hm,Yu2018-ns, Xu2020-jl}, medicine~\cite{Cai2020-as, Altae-Tran2017-uj, Prabhu2019-wi}, and more~\cite{Xing2019-ku, Bertinetto2016-cg, Brock2018-qy, Wei2019-hh}.
In the computer vision domain, a wide-range of approaches were proposed to tackle the few-shot image classification challenge.
Those include learning fast optimization schemes~\cite{Ravi2017-xu}, weight-imprinting~\cite{Qi2018-rw}, memory-augmented neural networks~\cite{Santoro2016-br}, casting the problem as stochastic process learning~\cite{Garnelo2018-pq}, or going about it from a Bayesian perspective\cite{Ortega2019-qi}.
This manuscript focuses on two major families of few-shot learning algorithms which we review below; for a more exhaustive account of few-shot and meta-learning, we refer the reader to the surveys of~\cite{Wang2020-fj} and~\cite{Vanschoren2018-mt}.

\paragraph{Gradient-Based Few-Shot Learning}
The main idea behind gradient-based few-shot learning algorithms is to discover a model whose parmeters can be adapted with just a few steps of gradient descent~\cite{Finn2018-fp}.
The representative algorithm of this family, MAML~\cite{Finn2017-gw}, does so by learning an initialization end-to-end while adaping all weights of the model.
Due to its flexible formulation, MAML was successfully applied to vision~\cite{Finn2017-gw}, robotics~\cite{Nagabandi2018-cz}, lifelong learning~\cite{Finn2019-yr}, and more~\cite{Zhou2020-ea}.
Variations of MAML improved upon its adaptation ability, reduce its computational footprint, or both.
Notable attempts to improve MAML's performance include meta-learning parameters dedicated to optimization, implicit~\cite{Lee2018-gw, Flennerhag2019-fw} or explicit~\cite{Li2017-li, Park2019-or, Behl2019-ki}, probabilistic regularization schemes~\cite{Grant2018-lh, Yoon2018-om}, as well as various training refinements~\cite{Antoniou2018-vl}.
To reduce the burden of computing second-order derivatives induced by the MAML objective, the authors suggested omitting those derivatives altogether at the cost of decreased performance.
When taking several gradient steps, one can leverage the implicit function theorem to more cheaply obtain the model updates~\cite{Rajeswaran2019-hx}.
Other options to mitigate the expense of second-order derivatives consists of conditioning the model on a latent embedding, and only updating that embedding during adaptation~\cite{Rusu2018-op, Rakelly2019-vk, Zintgraf2018-ot}.
In the same vein, the authors of ANIL~\cite{Raghu2019-ff} suggest that it is sufficient to update the very last layer of neural network to reap the benefits of MAML; that is, the model learns a feature extractor shared across all tasks and simply update the linear classifier a few times for each new task.

\paragraph{Metric-Based Few-Shot Learning}
Similar to ANIL, metric-based methods also share a feature extractor across tasks to extract high-dimensional embedding representations.
But, rather than adapting a linear classifier, they compute the distance of new instances' embeddings to the few-shot set of reference embeddings, akin to nearest neighbour classification.
Matching Networks~\cite{Vinyals2016-cm}, which was proposed to tackle the special case of one-shot learning, uses the (negative) cosine similarity as distance measure and relies on two different networks, one for the query and one for the reference embedding.
Prototypical Networks (ProtoNet)~\cite{Snell2017-kv} generalizes Matching Networks to the few-shot setting and measures similarity between embeddings with the Euclidean distance.
Some extensions to ProtoNet centered around learning and improving a specialized distance metric for a given task, typically by solving a convex optimization problem~\cite{Bertinetto2018-gn, Lee2019-cl, Zhang2020-ks}.
Others attempted to improve the embedding representations of a task~\cite{Oreshkin2018-yi, Rodriguez2020-vm, Ye2020-wf, Lichtenstein2020-py}.
Finally, ProtoMAML~\cite{Triantafillou2019-bj} combined MAML with ProtoNet such that the classification layer is initialized with the reference embedding values and the model is adapted for a few steps of gradient descent.

A recent line of work put a lot of those advances in question, and show that a simple baseline matches -- and often surpasses -- state-of-the-art performance in few-shot classification~\cite{Tian2020-gv, Wang2019-fc, Dhillon2019-mg, Chen2020-yd}.
Rather than learning by sampling many different tasks, they suggest to aggregate the classes from all tasks and to train a model via standard one-vs-rest multiclass classification.
At test time, the last linear layer of the model is removed to obtain a trained feature extractor, which can be used \`a la ProtoNet.
Directly~\cite{Tian2020-gv} or not, those results beg the question: \emph{do we still need to adapt our features for few-shot learning?}

\paragraph{Tasksets for Few-Shot Learning}
Our manuscript tackles this question when transfer to new tasks is especially challenging.
In this context, few-shot classification tasksets can be broadly categorized in two groups: those that leverage additional human information and those that don't.
In the latter case, the standard approach consists in taking an existing classification dataset and randomly partitioning classes in train and test sets.
Typical examples include CIFAR-FS~\cite{Bertinetto2018-gn}, mini-ImageNet~\cite{Vinyals2016-cm, Ravi2017-xu}, TCGA~\cite{Samiei2019-rx}, MultiMNIST~\cite{Sabour2017-ml, Sener2018-uc}, CU Birds~\cite{Welinder2010-cj}, FGVC Planes~\cite{Maji2013-ct} and Fungi~\cite{Sulc2020-vh, Schroeder2018-fu}, VGG Flowers~\cite{Nilsback2006-ye}, and Omniglot~\cite{Lake2015-uv, Lake2019-ph}.
Tasksets that do take advantage of human knowledge (usually in the form of semantic class relationships) attempt to minimize \emph{information overlap} to challenge transfer.
To the best of our knowledge, there are two such tasksets: FC100~\cite{Oreshkin2018-yi} and tiered-ImageNet~\cite{Ren2018-hk}.
The former is built on CIFAR100~\cite{Krizhevsky2009-mj} and leverages its superclass structure, while tiered-Imagenet is a subset of ImageNet-1k~\cite{Russakovsky2015-he} and uses the WordNet~\cite{Miller1990-zm} database.

Building on those datasets (and some more) is Meta-Dataset~\cite{Triantafillou2019-bj}: a large collection of tasksets aimed at mimicking transfer in real-world scenarios.
For example, one benchmarking scenario consists of training a model on ImageNet-1k and evaluating its performance on the remaining 9 tasksets.
Similar in spirit is VTAB~\cite{Zhai2019-wt}, which also consists of a large collection of synthetic and natural images aimed at evaluating representation learning and transfer.

In contrast to prior work, we study the question raised by Tian \emph{et al.}~\cite{Tian2020-gv}: is a good embedding sufficient for few-shot learning?
We approach this question from the standpoint where transfer is particularly challenging.
Since training gradient-based methods on large benchmarks such as Meta-Datasets and VTAB is prohibitively expensive, we devise a method to automatically generate tasksets with explicit control over transfer difficulty.


\section{Background and Notation}
\label{sBackground}

We now present the few-shot learning setting, introduce notation, and review the few-shot algorithms of interest to the remainder of this paper.

\paragraph{Datasets, Tasksets, and Tasks}
We designate by \emph{dataset} a set of input-class pairs $(\x, y)$, where the dependent variable $y$ takes values from a finite set of classes $y_1, \dots, y_M$.
This dataset induces a classification \emph{task}, which consists of finding a predictive function mapping $\x$ to $y$.
One common approach to construct a set of such tasks, \ie a \emph{taskset}, is to first sample a subset of $N$ classes, and then sample $K$ input-class pairs for each of those classes.
This setting is usually referred to as $N$-ways $K$-shots.

As we often wish to evaluate the transferability of few-shot models from one taskset to another, we first partition the $M$ classes of the dataset into train and test splits and construct tasksets from those splits.
As mentioned in Section~\ref{sIntroduction}, partitioning can either be random, in which case train and test classes come from the same distribution, or it can leverage additionaal human knowledge.
This additional information typically defines a hierarchy over the classes (\eg semantic class relationships), which can be used for partitioning.
For example, the 100 classes of CIFAR100 are grouped into 20 superclasses, and FC100 uses 12 of those superclasses (60 classes) for training and 4 (20 classes) for validation and testing, each.
Once tasksets are defined, the goal of a few-shot learning algorithm is to find a model of the train tasks that generalizes to the test tasks.

To achieve this goal, we associate to each task $\tau$ a loss $\mathcal{L}_\tau$ and a parameterized model $p_\tau(y \mid \x)$.
Without loss of generality, we write this model as the composition of a linear layer $w$ and a feature extractor $\phi$ (\eg a neural network).
Then, finding a maximum likelihood solution for $\phi$ and $w$ boils down to minimizing the average task loss $\Exp[\tau]{\mathcal{L}_\tau(\phi, w)}$ over the training tasks.

\paragraph{Algorithms}
Gradient-based methods, and MAML in particular, compute the likelihood after adapting $\phi$ and $w$ for one (or a few) gradient step.
For the $i$th class, this likelihood takes the form:
\begin{align*}
    \tag{MAML}
    p_\tau (y=y_i \mid \x) &= \softmax( \phi^\prime (\x)^\top w_i^\prime) \\
    \text{s.t.} \quad w^\prime &= w - \alpha \nabla_{w} \mathcal{L}_\tau \\
                   \phi^\prime &= \phi - \alpha \nabla_\phi \mathcal{L}_\tau
\end{align*}
where $\alpha$ denotes the adaptation learning rate, and $w_i^\prime$ is the $i$th column of the adapted linear layer.
In the adaptation step, the gradient of the loss is computed over the set of few-shot examples; differentiating it requires Hessian-gradient products which makes MAML an expensive method for large feature extractor $\phi$.
For this reason, MAML is rarely used with architectures larger than the standard 4-layer CNN.
To scale up MAML to the 12-layer ResNet of our experiments, we implement a data-parallel version of MAML where different GPUs are responsible for different subsets of the few-shot inputs.
Refer to the Supplementary Material for more details.

To alleviate this computational burden, the authors of ANIL suggest that adapting $w$ is sufficient to claim many of MAML's benefits:
\begin{align*}
    \tag{ANIL}
    p_\tau (y=y_i \mid \x) &= \softmax( \phi (\x)^\top w_i^\prime) \\
    \text{s.t.} \quad w^\prime &= w - \alpha \nabla_{w} \mathcal{L}_\tau.
\end{align*}
Contrasting with MAML, the difference lies in the feature extractor $\phi$ being shared -- \emph{not} adapted -- across tasks.
In that sense ANIL resembles metric-based based methods, despite its original motivation of approximating MAML.
Since $\phi$ is responsible for the bulk of the computation, ANIL becomes increasingly more efficient as the feature extractor grows in size.\footnote{In our experiments, we saw speed-ups as large as 9.25x over MAML.}

A representative metric-based algorithm, ProtoNet also shares $\phi$ across tasks and replaces the final linear classifier by a nearest neighbour one.
It measures this nearest neighbour distance between the query embedding $\phi(\x)$ and the $i$th class embedding $\phi_i$:
\begin{align*}
    \tag{ProtoNet}
    p_\tau (y=y_i \mid \x) &= \softmax\left( - d(\phi (\x), \phi_i)\right)
\end{align*}
where $\phi_i = \Exp[\x \sim p_\tau(\cdot \mid y_i)]{\phi (\x)}$ is the average embedding of the few samples with class $y_i$, and $d(\cdot, \cdot)$ is a distance function -- common choices include the Euclidean norm or the (negative) cosine similarity.

Since the ProtoNet classifier is nonparametric, it can be used with any feature extractor $\phi$.
Thus, a simple yet effective baseline, \emph{Multiclass}, consists of collapsing all train tasks into a single large tasks and learning a feature extractor via standard one-vs-rest multiclass classification.
At test-time, we can readily use this learned feature extractor with the ProtoNet classifier to solve unseen tasks.


\section{Taskset Generation: Easy or Hard?}
\label{sMethod}

This section describes \ourmethod, a method for generating tasksets without requiring additional human knowledge, and which provides fine-grained control over the transfer difficulty between train and test tasks.

Our goal is to partition classes into train and test sets, such that we can control the difficulty of transferring a model trained on one set to the other.
At a high-level, \ourmethod finds those class partitions based on their \emph{class embedding} $\phi_1, \dots, \phi_M$.
Each class embedding $\phi_i$ is obtained by averaging the embedding of all samples from its corresponding class $y_i$:
\begin{equation*}
    \phi_i = \Exp[\x \sim p(\cdot \mid y = y_i)]{\phi(\x)},
\end{equation*}
were $\phi$ is a pretrained feature extractor.
In practice, we find that pretraining $\phi$ on the dataset to be partitioned leads to satisfying results, but simpler approaches might also work. (\eg an off-the-shelf model pretrained on ImageNet)

The key feature of \ourmethod is that train and test clusters are encouraged to stay at a \emph{prespecified distance of each other} in the class embedding space.
Intuitively, the farther and tighter the test set, the more difficult it will be for few-shot methods to discriminate between those classes.
On the other hand, a test task is easily solved if its classes are dissimilar (\ie pooly clustered) and each class is fairly similar to classes seen during training (\ie train and test clusters are adjacent).

To formalize this intuition, we pretend that the class embeddings were sampled from the mixture of two distributions $\ptrain(y)$ and $\ptest(y)$.
We model $\ptrain$ and $\ptest$ as multinomials such that their density for class $y_i$ depends on its distance to the distribution's centroid:
\begin{align*}
    \ptrain(y = y_i) &= \softmax (\vert\vert \phi_i - \mutrain \vert \vert^2), \quad\text{and} \\
    \ptest(y = y_i) &= \softmax (\vert\vert \phi_i - \mutest \vert \vert^2),
\end{align*}
where the centroids $\mutrain$ and $\mutest$ are learnable parameters.
Maximizing the log-likelihood of the mixture $\frac1{2} (\ptrain + \ptest)$ is identical to a soft K-means objective if $\ptrain$ and $\ptest$ were Gaussians~\cite[Chapter 22]{MacKay2002-la}.
In practice, we preferred multinomials because Gaussians led to numerical instabilities when optimizing the penalized objective (see below) by gradient descent.

We can incentivize $\ptrain$ and $\ptest$ to lie at a given distance $R \geq 0$ by including a penalty $(D(\ptrain \vert\vert \ptest) - R)^2$ based on our choice of statistical divergence $D$.
Combining this penalty with the mixture's log-likelihood yields our final penalized clustering objective:
\begin{align*}
    J = &- \sum_{i=0}^M \log \frac1{2} \left( \ptrain(y_i) + \ptest(y_i)\right) \\
        &+ \lambda \left( D(\ptrain \vert \vert \ptest) - R \right)^2,
\end{align*}
where $\lambda \geq 0$ balances the value of clustering versus the penalty.
Assuming train and test classes tightly cluster, controlling the transfer difficulty amounts to controlling the distance $D(\ptrain \vert\vert \ptest)$ between the two distributions, which we can achieve by adjusting $R$.

\paragraph{Train, Validation, and Test Assignments}
When $J$ is minimized with respect to $\mutrain$ and $\mutest$, we get a solution we can use to partition $y_1, \dots, y_M$.
One approach is to assign class $y_i$ to the train set if the ratio $\frac{\ptrain(y_i)}{\ptest(y_i)}$ is greater than 1, and to the test set if less.\footnote{Ties are broken at random, so that if $D(\ptrain \vert \vert \ptest) = 0$ assignments are also random.}

Partitioning according to this decision rule is theoretically sound but might lead to slightly degenerate solutions in the context of few-shot learning.
For example, we observed instances where only 3 classes were assigned to the test set, which prevents evaluation in the 5-ways setting.
It also doesn’t prescribe how to devise a validation set from the test distribution.
To resolve those issues, we select the top 60\% scoring classes and assign them to the train set.
The remaining 40\% classes are split among validation and test sets in turn, according to their score: the least scoring class is assigned to the test set, the second-least scoring to the validation, etc...
Thus, validation and test classes retain roughly equal probability under $\ptest$ and the resulting tasks are of similar difficulty.


\section{Experiments}
\label{sExperiments}

Our experiments focus on three aspects.
First, we empirically validate our proposed method to control the information overlap between train and test classes.
To provide an intuitive picture of our method, we also perform an ablation study on the different measures of information and visualize low-dimensional projections of the resulting partitions.
Second, we compare the partitions obtained from our method to the ones obtained using semantic class relationships.
Our results suggest that both types of partitions are significantly different; moreover, they confirm that our method is capable of generating partitions that are more challenging than the ones resting on semantics.
Finally, we leverage our partitions to compare gradient-based and metric-based classification methods.
Those experiments indicate that metric-based methods -- which were recently thought to be sufficient for few-shot learning -- tend to underperform in challenging transfer settings when compared to their gradient-based counterpart.

For additional experimental details, including the exact partitions for all tasksets, please see the Supplementary Material.

\subsection{Setup}
Our experiments build on architectures and datasets widely used in the computer vision and few-shot learning literature.
We denote by CNN4 the 4-layer CNN with 64 hidden described in~\cite{Snell2017-kv}, which we use for few-shot learning experiments on FC100, CIFAR-FS, EMNIST, and LFW10.
(The latter two datasets are described below.)
On mini-ImageNet and tiered-ImageNet, we use the 12-layer residual network (ResNet12) described in~\cite{Mishra2017-vm}, and add the DropBlock layers proposed in~\cite{Lee2019-cl} for regularization.
(But unlike Lee~\emph{et al.} we keep the final average pooling layers.)
For experiments involving Imagenet-1k -- and including Birds, Planes, Flowers, Fungi --  we use 3 architectures: a 121-layer DenseNet (DenseNet121;~\cite{Huang2016-qc}), a 18-layer residual network (ResNet18;~\cite{He2015-qd}), and a GoogLeNet (GoogLeNet;~\cite{Szegedy2014-rn}).
For all models, we define the feature extractor to be the architecture up to the last fully-connected layer.

All tasksets built with \ourmethod follow an identical recipe.
We pretrain $\phi$ on the same data used to compute the class embeddings $\phi_i$.
We minimize $J$ by gradient descent with $\lambda = 1$, and, unless specified otherwise, use the symmetrized version of the KL divergence for $D$.

With this manuscript, we also contribute tasksets for two datasets previously not used in the few-shot learning setting: EMNIST and LFW10.

\paragraph{EMNIST}
The Extended MNIST dataset (EMNIST;~\cite{Cohen2017-vw}) is a variant of the MNIST dataset consisting of 814,255 grayscale images of 62 handwritten characters (digits, lowercase and uppercase alphabet).
Each image contains a single character scaled to 28x28 pixels.
Using \ourmethod, we partition its classes in 37 characters for training, 12 for validation, and 13 for testing, filling a niche in the few-shot dataset landscape: EMNIST is lightweight (roughly 610Mb) with few training classes, but plenty of data ($>$10k) available per class.

\paragraph{LFW10}
We bootstrap LFW10 from the \emph{Labelled Faces in the Wild} dataset~\cite{Huang2014-hw}, which contains 13,233 pictures of 5,749 famous personalities. 
Of those 5k personalities, we select all 158 that have at least 10 images in the dataset -- fewer would constrain the few-shot learning setup -- and partition them in tasksets of 94 train, 32 validation, and 32 test classes.
Each image is rescaled to 62x47 pixels with RGB colors.
LFW10 is an example of dataset for which collecting class relationships is difficult if not impossible, as it requires putting a semantic hierarchy over each individual in the dataset.

\subsection{Controlling Information Overlap}\label{sControllingInformationOverlap}


\begin{figure*}[ht]
    \begin{center}
        \includegraphics[height=0.205\linewidth]{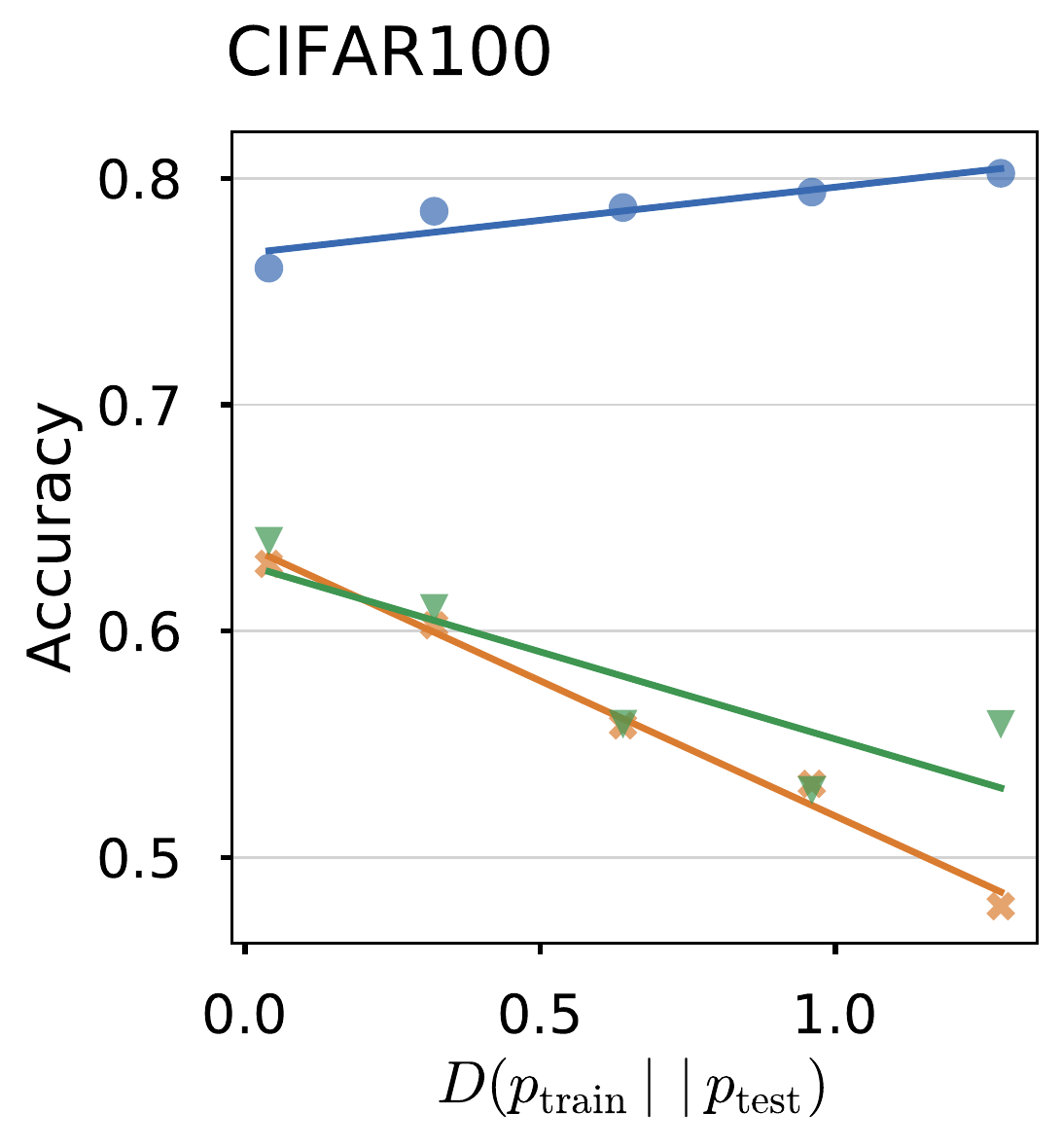}
        \includegraphics[height=0.205\linewidth]{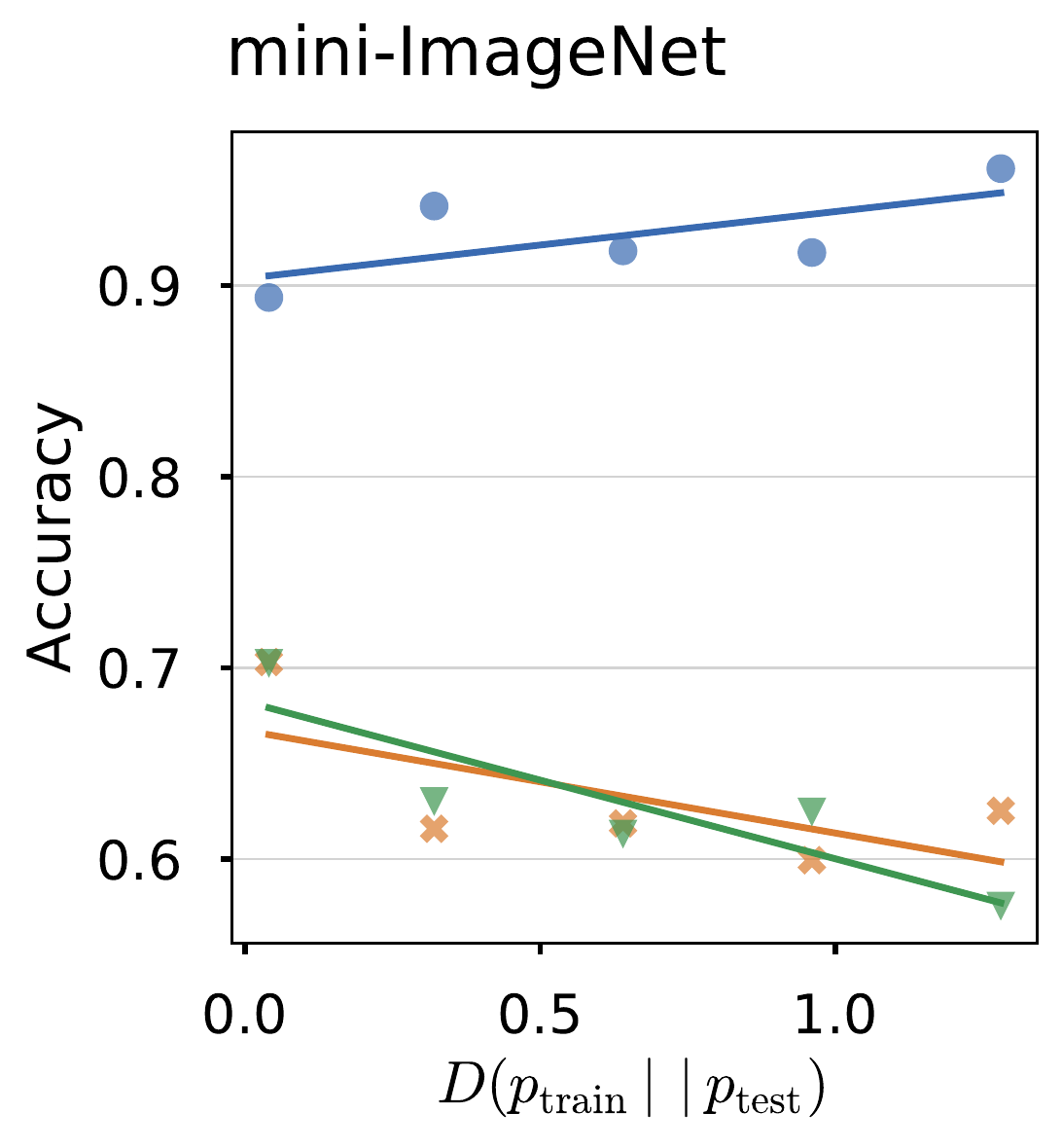}
        \includegraphics[height=0.205\linewidth]{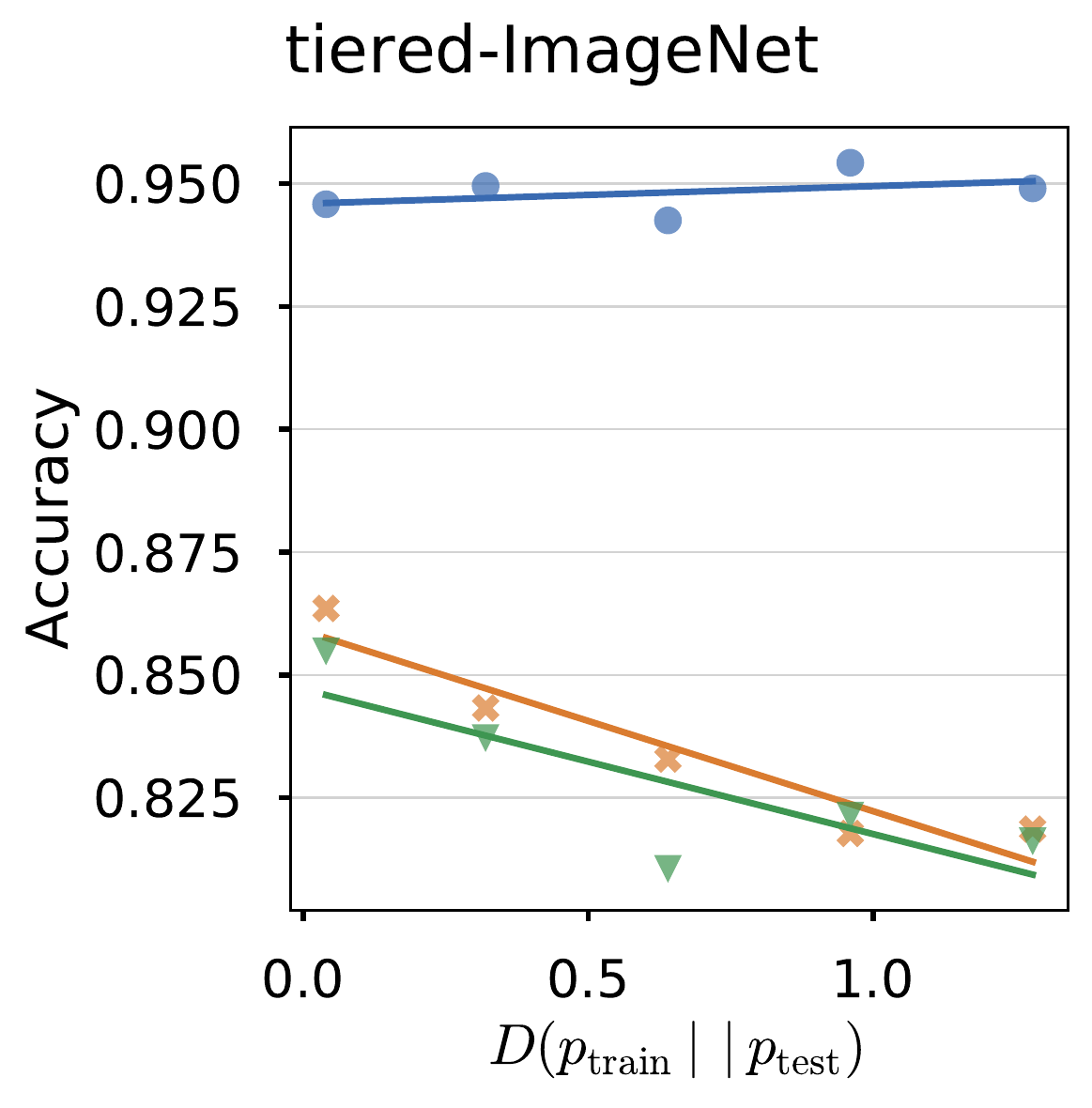}
        \includegraphics[height=0.205\linewidth]{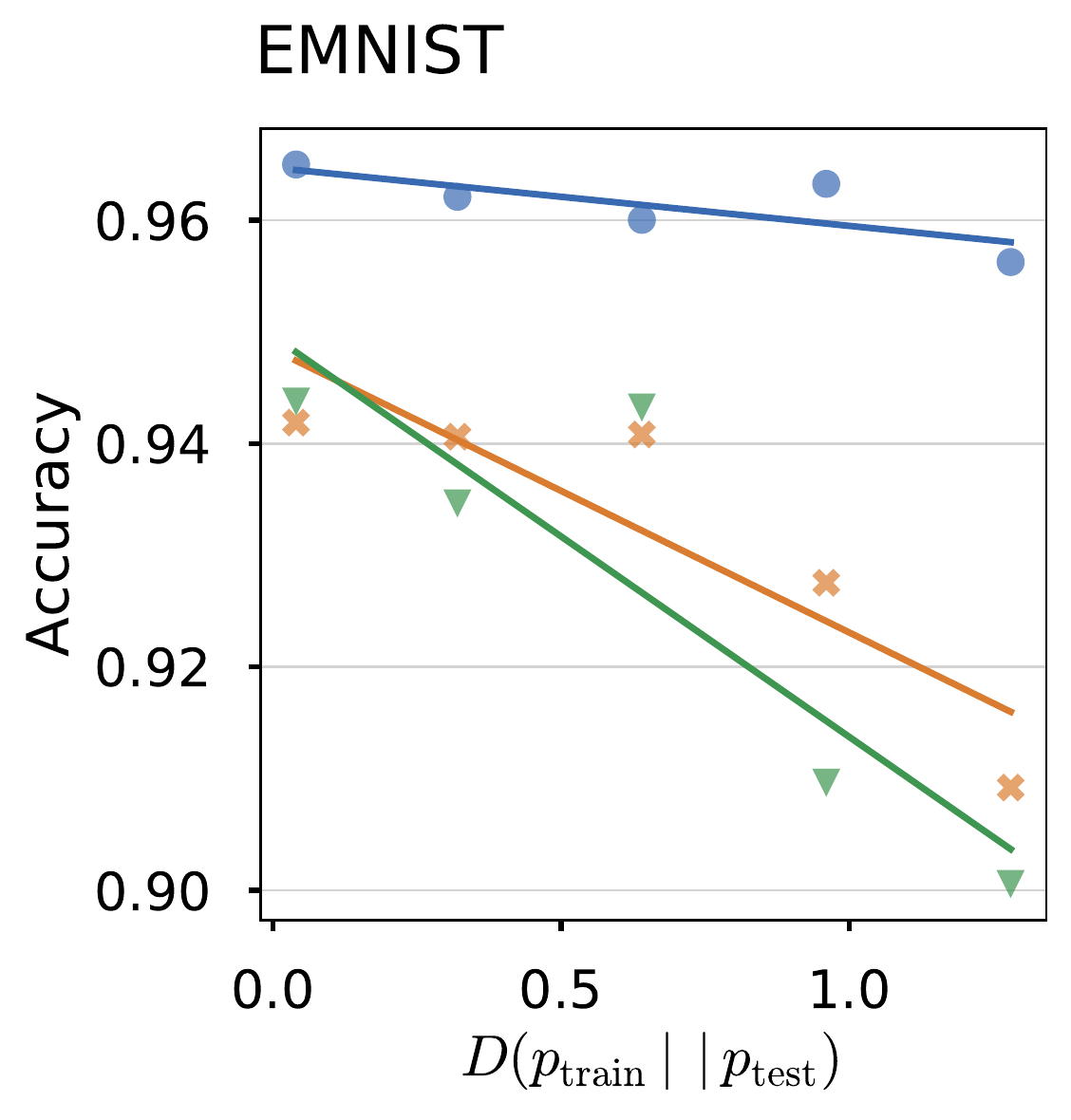}
        \includegraphics[height=0.205\linewidth]{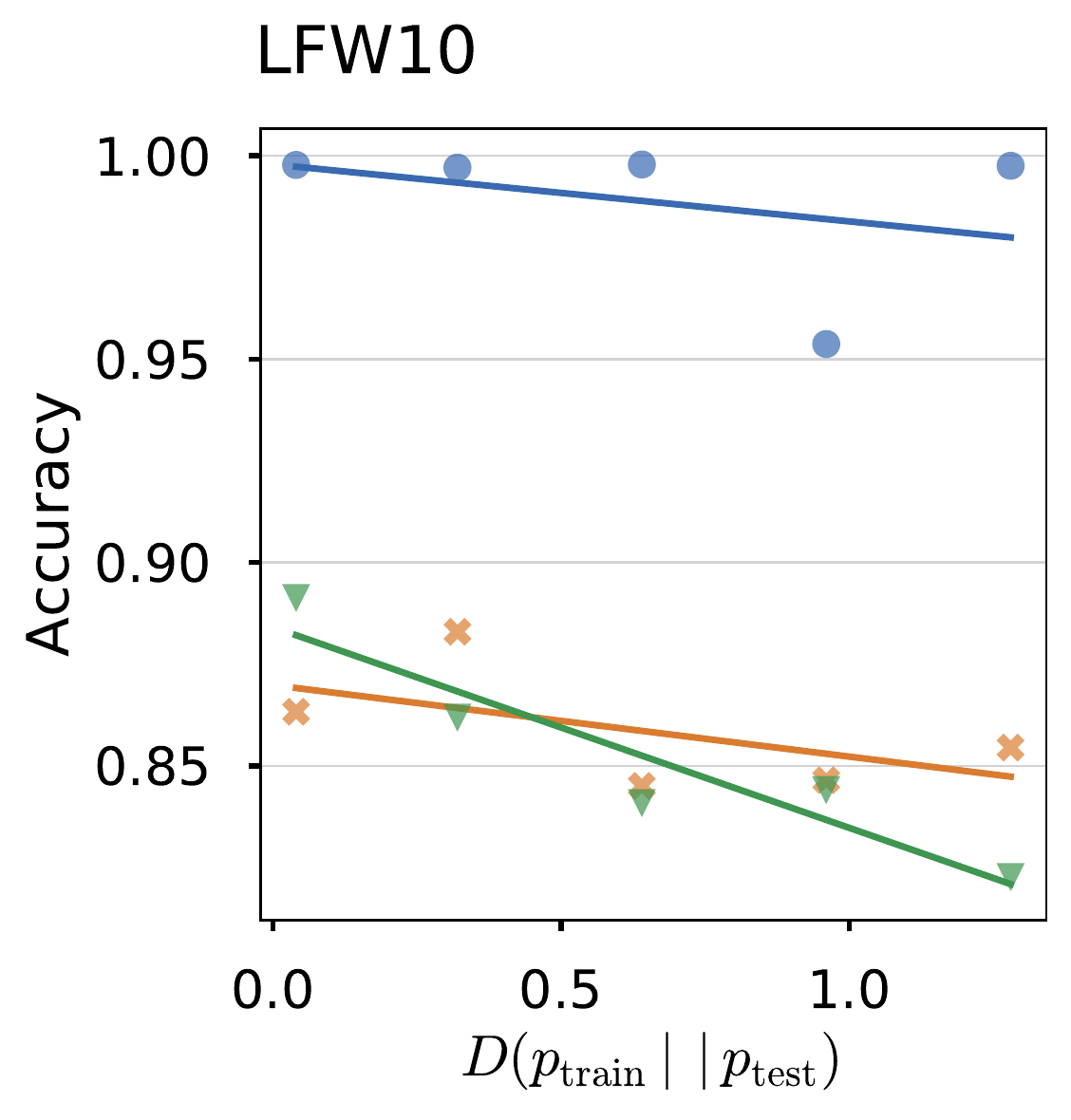}
        \\
        \vspace{0.2em}
        \hspace{3.0em}\includegraphics[width=0.40\linewidth]{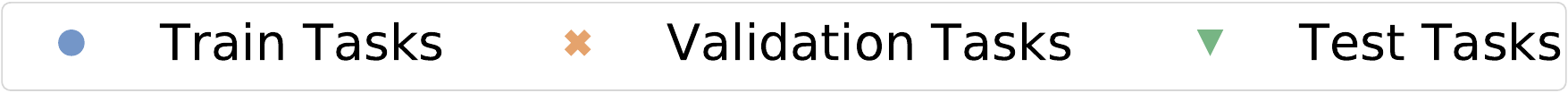}
    \end{center}
    \vspace{-1em}
    \caption{
        \small
        Accuracy of a Multiclass-trained netwok as we increase the divergence between train and test class distributions.
        As the divergence increases, accuracy drops suggesting that the divergence can be used to generate tasksets of varying difficulty.
    }
    \label{fMIAcc}
\end{figure*}

We first test whether our method is effective in producing tasksets of varying difficulty.
To that end, we use \ourmethod to generate tasksets of increasing difficulty on the classes of 5 datasets: CIFAR100, mini-ImageNet, tiered-ImageNet, LFW10, and EMNIST.
First, we train a convolutional network over the entire set of classes, using the standard cross-entropy minimization.
We use a 121-layer DenseNet for CIFAR100, mini- and tiered-ImageNet dataset, and the CNN4 for LFW10 and EMNIST.
We then remove the last fully-connected layer of the network, and use this feature extractor to compute the mean embedding of each class.
Finally, we create train, validation, and test class partitions for various target divergence between train and test classes.

We measure transfer difficulty with the 5-ways 5-shots classification accuracy of a model trained with Multiclass on the train taskset only.
We use the ResNet12 for ImageNet-based tasks, and the CNN4 for the others.
Figure~\ref{fMIAcc} reports train, validation, and test accuracies as a function of the divergence for all datasets.
Across all datasets test accuracy decreases as the divergence between train and test class distributions increases.
Thus, we conclude that our method is effective in finding partitions of desired transfer difficulty.
We also observe that validation and test accuracies similarly challenge the transfer ability of the Multiclass model, which is expected as both sets of classes come from the same distribution.


\begin{figure}[h]
    \vspace{-1em}
    \begin{center}
        \includegraphics[width=0.43\linewidth]{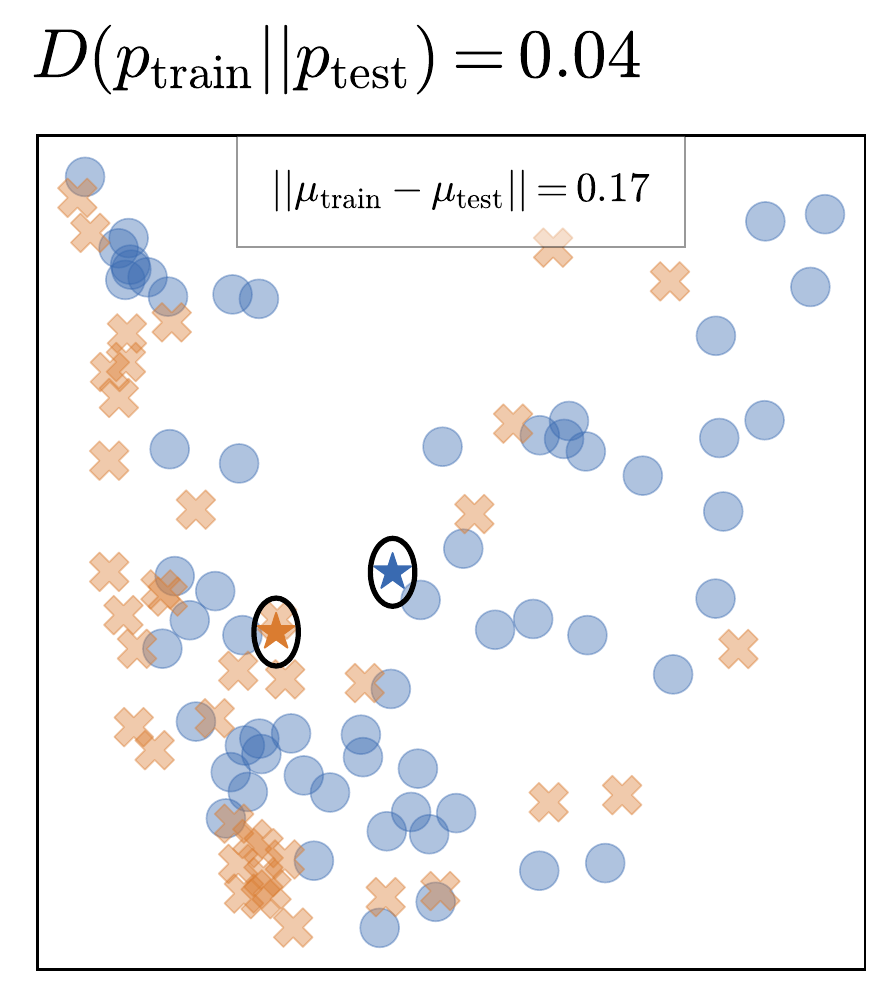}
        \includegraphics[width=0.43\linewidth]{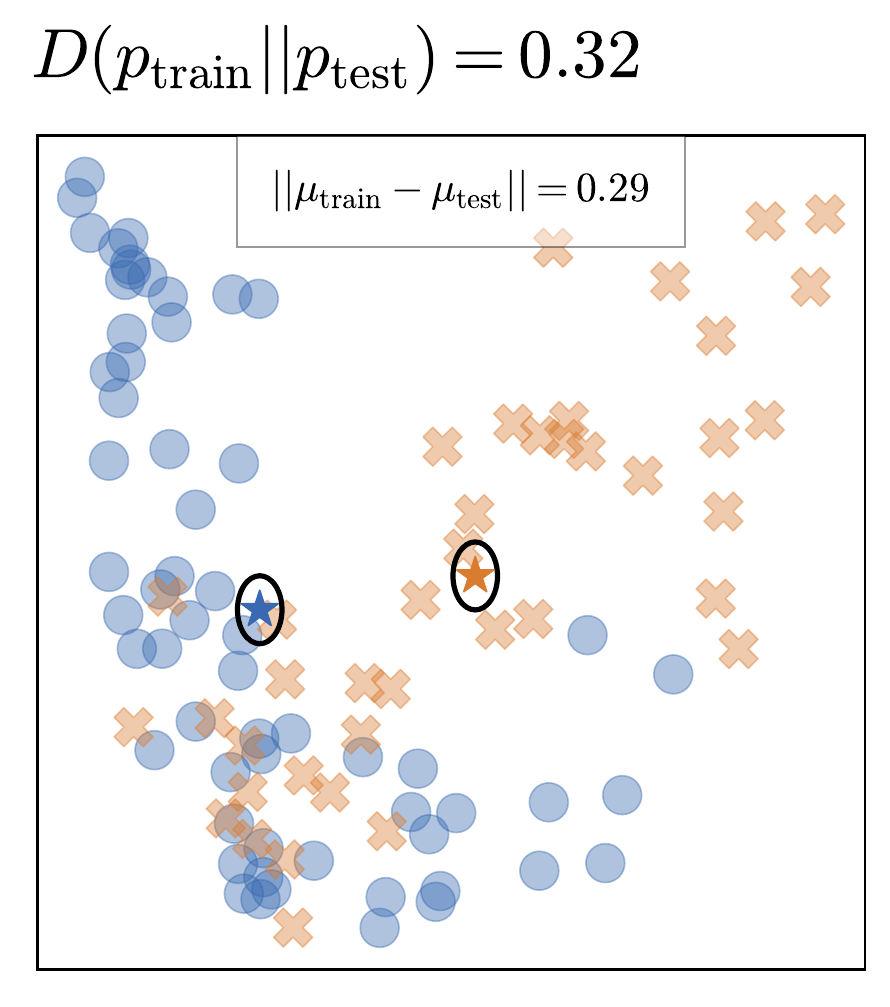} \\
        \includegraphics[width=0.43\linewidth]{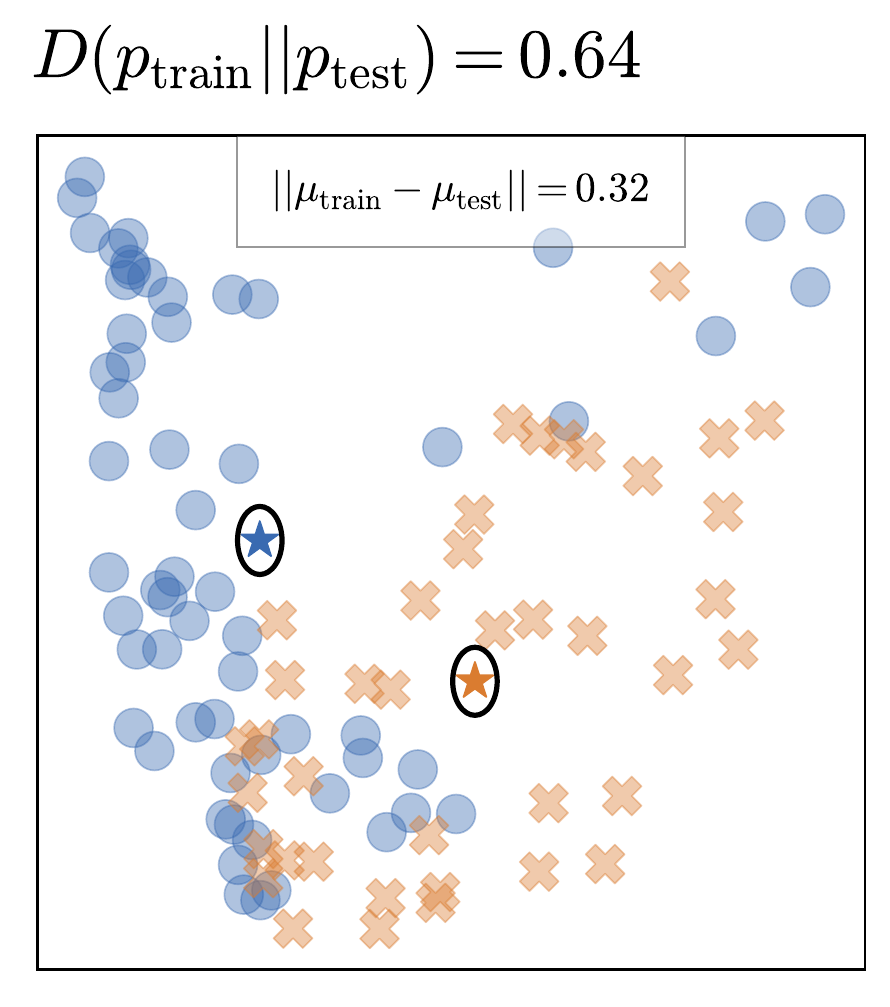}
        \includegraphics[width=0.43\linewidth]{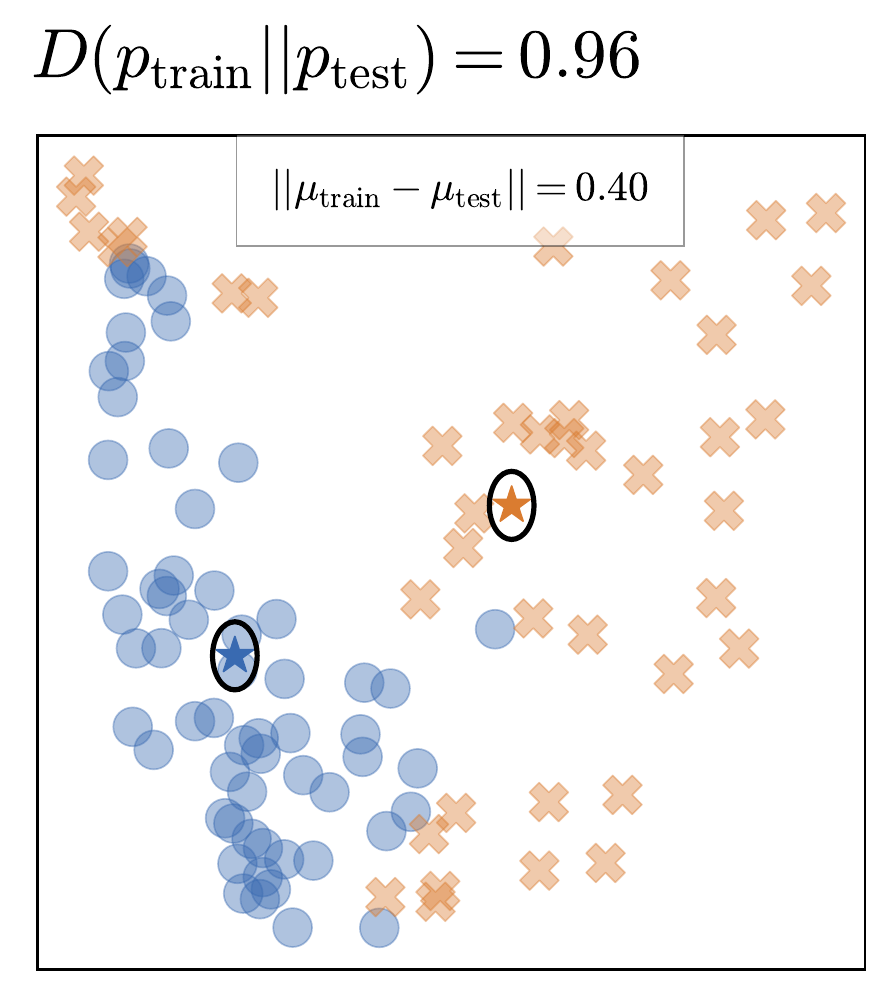} \\
        \includegraphics[width=1.00\linewidth]{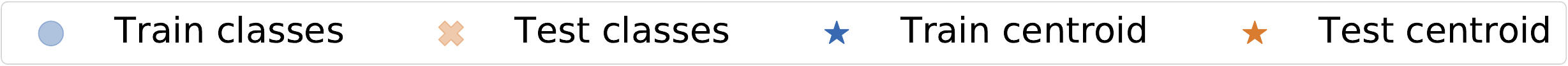}
    \end{center}
    \vspace{-1em}
    \caption{
        \small
        Low-dimensional projections of class embeddings and taskset centroids.
        As we increase the divergence penalty, the centroids spread further apart.
    }
    \label{fPCAClustering}
    \vspace{-1em}
\end{figure}

\paragraph{PCA Visualizations}
We perform two more ablative studies to provide further understanding of \ourmethod.
In Figure~\ref{fPCAClustering}, we plot the 2-dimensional PCA projection of the CIFAR100 mean class embeddings, together with their assignments and the assigned centroid.
We observe that centroids separate further as the divergence increases from 0.04 to 1.28 -- the greater the divergence between $p_\text{train}$ and $p_\text{test}$, the further $\mu_\text{train}$ and $\mu_\text{test}$.

\paragraph{Comparing Divergences}
Although intuitive, this description is not sufficient to explain why \ourmethod works because it lacks a measure of how difficult it is to discriminate between the classes of a task.
To show this, we compare different methods for measuring the distance of train and test class distributions between ImageNet-1k and 4 downstream datasets.
(Birds, Planes, Fungi, and Flowers)
We take a 121-layer DenseNet pretrained on ImageNet-1k, and compute the mean class embedding for all 1,000 classes as well as the ImageNet centroid.
Then, for each of the downstream dataset, we sample 100 tasks, each consisting of 5 classes and 5 samples.
Finally, we compute the distance between the train distribution (defined by the ImageNet centroid) and the task distribution (defined by the centroid of the task) over all 1,005 classes.


\begin{table}[h]
    \vspace{0em}
    \centering
    \caption{
        \small
        Correlation between divergence and accuracy for different choice of divergence $D$.
        Measuring the Euclidean distance between centroids performs worst, while the symmetrized KL divergence (used in our other experiments) is best.
    }
    \label{tMeasureOverlap}
    \setlength{\tabcolsep}{3pt}
    {\small
    \begin{tabular}{l cccc}
        \toprule

                             & Birds     & Planes    & Fungi     & Flowers \\
        \midrule
        Euclidean Distance   & -0.50     & -0.08     & -0.17     & -0.33 \\
        Wasserstein-2        & -0.57     & -0.24     & -0.23     & -0.30 \\
        Kullback-Leibler     & -0.63     & -0.41     & \h{-0.51} & -0.29 \\
        Symmetrized KL       & \h{-0.73} & \h{-0.43} & -0.49     & \h{-0.43} \\

        \bottomrule
    \end{tabular}
    }
    \vspace{0em}
\end{table}

Table~\ref{tMeasureOverlap} reports the Pearson’s correlation coefficient between those distance measurements and the task accuracies for all datasets.
The symmetrized KL, which we use in our remaining experiments, performs best while naively measuring the Euclidean distance between centroids (Euclidan) performs worse.
As a final remark, let us observe that correlations have fairly low magnitudes, indicating that our method is ill-suited to accurately measure individual task difficulty.
Assessing task similarity and difficulty is an open research question; we refer the reader to related litterature~\cite{Achille2019-hu, Fifty2020-my, Dhillon2019-mg, Wu2019-kw} for more details.

\subsection{Comparing Semantic vs Embedding Clusters}
We continue the study of our method and zero-in on the difference between class embeddings and semantic to encode class similarities.

\paragraph{Comparison to WordNet}
The first question we ask is whether class embeddings capture information similar to semantics.
To that end, we compare hierarchies created by clustering class embeddings to WordNet, the hierarchy over ImageNet-1k classes induced by class semantics.

We construct three sets of ImageNet-1k embeddings with three different pretrained architectures.
For each set, we obtain a tree of their classes via Ward clustering.
To compare those trees against the WordNet graph of classes, we define the \emph{hop} distance -- the average difference of distances between two classes $a, b$:
\begin{equation*}
    \sum_{a, b} \vert d_\text{Clustering}(a, b) - d_\text{WordNet}(a, b) \vert,
\end{equation*}
where $d_\text{Clustering}, d_\text{WordNet}$ are the normalized minimum number of nodes separating $a$ to $b$ in the clustering tree and WordNet graph, respectively.
We use such a cumbersome metric because WordNet contains cycles as well as several vertices of degree 2; intuitively, this metric can be understood as the average difference in the number ``hops’’ necessary to reach $b$ starting from $a$ between the two graphs.


\begin{table}[h]
\vspace{0em}
\centering
\caption{
    \small
    Average hop distance between WordNet and hierarchies created from Imagenet-1k embeddings (via hierarchical clustering).
    Regardless of the network architecture, trees constructed from class embeddings are more similar to each other than WordNet, indicating that class partitioning relies on attributes different from semantic relationships.
}
\label{tHopDistances}
\setlength{\tabcolsep}{3pt}
{\small
\begin{tabular}{l ccc}
\addlinespace
\toprule

                        & ResNet18 & DenseNet121 & GoogLeNet \\
\midrule
ResNet18                & 0.0      & 1.99        & 2.18 \\
DenseNet121             & 1.99     & 0.0         & 2.26 \\
GoogLeNet               & 2.18     & 2.26        & 0.0 \\
\midrule
WordNet                 & \h{8.60} & \h{8.35}    & \h{8.59} \\


\bottomrule
\end{tabular}
}
\vspace{0em}
\end{table}

Table~\ref{tHopDistances} reports the hop distance between WordNet and clustering trees, when the class embeddings are computed using DenseNet121, GoogLeNet (embedding sizes 104) and ResNet18. (embedding size 512)
Regardless of embedding size or network architecture, the clustering structures differ from the WordNet structure by a factor of 4, indicating that class embeddings encode attributes significantly different from class semantics.

Anecdotally, we manually inspect the trees generated from embeddings in an attempt to shed some light on what those poperties might be.
We find that the mean class embeddings tended to encode visual properties.
For example, the \emph{traffic light} and \emph{theatre} classes are clustered close-by (3 hops) due to both containing many red pixels.
On the other hand, the Orangutan and Macaque classes (fairly close semantic-wise) are separated by 9 hops — likely due to the difference in color and texture of both animal’s furs.

\paragraph{Comparison across Methods}
Having established that \ourmethod can generate tasksets different from those built on semantics, we answer the next natural question: \emph{How do those tasksets compare to existing tasksets?}
Accordingly, we select the easy ($D=0.04$) and hard ($D=0.96$) tasksets generated with \ourmethod and train networks with 4 popular few-shot methods (MAML, ANIL, ProtoNet, Multiclass) on the 5-ways 1-shot and 5-ways 5-shots setting on the tasksets of CIFAR100, mini-ImageNet, and tiered-ImageNet.
To ensure fair comparison, we do not augment the data and use the exact same architecture for all methods within a dataset.
(CNN4 for CIFAR100, ResNet12 for mini/tiered-ImageNet.%
\footnote{Training a ResNet12 with MAML on tiered-ImageNet was \emph{still} too computationally expensive, despite our data-parallel implementation.})
Each (method, taskset) pair is tuned independently, using a logarithmic-spaced grid-search.
For MAML and ANIL, we measure accuracy after 5 adaptation steps.


\begin{table*}[h]
    \vspace{-1em}
    \renewcommand{\arrayrulewidth}{1.3}
    \centering
    \caption{
        \small
        Comparing classification accuracy of different tasksets for a same dataset across popular few-shot learning methods.
        Our proposed method, \ourmethod, is capable of generating simple tasksets (close to random partitioning) as well as challenging ones.
        In particular, it is often more challenging than tasksets built with class semantics (denoted with a~$^\dagger$), but unlike those it does not require additional information.
        \h{Bolded} results indicate most challenging taskset for a given method.
    }
    \label{tAccuracyComparison}
    \setlength{\tabcolsep}{3pt}
    \vspace{-1em}
    {\small
    \begin{tabular}{@{}lll cccc c  ccccc@{}}
        \addlinespace
        \toprule
        & & & \multicolumn{4}{c}{5-ways 1-shot} &        & \multicolumn{4}{c}{5-ways 5-shots}\\
        \cmidrule{4-7} \cmidrule{9-12}

        Dataset            & Tasksets           & Backbone & MAML        & ANIL        & ProtoNet    & Multiclass  &  & MAML        & ANIL        & ProtoNet    & Multiclass \\
        \midrule
        CIFAR100           & CIFAR-FS           & CNN4     & 56.96\%     & 54.47\%     & 54.97\%     & 54.82\%     &  & 72.99\%     & 69.44\%     & \e{72.00\%} & \e{68.83\%} \\
                           & FC100$^\dagger$    & CNN4     & 36.99\%     & 35.54\%     & 36.25\%     & 36.59\%     &  & \h{51.48\%} & 50.12\%     & 51.16\%     & \h{51.22\%} \\
                           & Random             & CNN4     & 56.66\%     & 51.43\%     & 51.25\%     & 52.39\%     &  & \e{73.16\%} & \e{69.81\%} & 71.12\%     & 69.24\% \\
                           & Easy (ours)        & CNN4     & 48.35\%     & 46.03\%     & 46.68\%     & 46.92\%     &  & 65.19\%     & 59.43\%     & 63.66\%     & 63.97\% \\
                           & Hard (ours)        & CNN4     & \h{35.86\%} & \h{33.55\%} & \h{35.59\%} & \h{35.44\%} &  & 55.73\%     & \h{49.27\%} & \h{50.46\%} & 53.02\% \\
        \midrule
        mini-ImageNet      & Original           & ResNet12 & 58.80\%     & 55.02\%     & 56.68\%     & 57.12\%     &  & 72.56\%     & \e{64.74\%} & \e{71.05\%} & 71.88\% \\
                           & Random             & ResNet12 & 53.12\%     & 50.34\%     & 51.58\%     & 51.88\%     &  & 72.25\%     & 62.77\%     & 69.70\%     & \e{72.03\%} \\
                           & Easy (ours)        & ResNet12 & 53.75\%     & 51.22\%     & 52.25\%     & 52.53\%     &  & 67.92\%     & 61.98\%     & 66.64\%     & 70.27\% \\
                           & Hard (ours)        & ResNet12 & \h{44.87\%} & \h{42.54\%} & \h{41.83\%} & \h{44.62\%} &  & \h{62.25\%} & \h{54.42\%} & \h{60.36\%} & \h{61.86\%} \\
        \midrule
        tiered-ImageNet    & Original$^\dagger$ & ResNet12 & n/a         & 56.99\%     & \h{61.59\%} & 66.75\%     &  & n/a         & 74.81\%     & 80.02\%     & 82.53\% \\
                           & Random             & ResNet12 & n/a         & 62.69\%     & 69.35\%     & 70.89\%     &  & n/a         & \e{80.60\%} & \e{85.46\%} & \e{86.98\%} \\
                           & Easy (ours)        & ResNet12 & n/a         & 59.53\%     & 64.39\%     & 68.16\%     &  & n/a         & 76.57\%     & 84.33\%     & 85.50\% \\
                           & Hard (ours)        & ResNet12 & n/a         & \h{56.19\%} & 62.91\%     & \h{65.48\%} &  & n/a         & \h{73.24\%} & \h{78.64\%} & \h{81.90\%} \\

        \bottomrule
    \end{tabular}
    }
    \vspace{-1em}
\end{table*}

Table~\ref{tAccuracyComparison} reports the test accuracy obtained at the best validation iteration.
In almost all scenarios, \ourmethod is able to generate tasksets that are as — or even more — challenging as the tasksets constructed from semantic relationships.
This makes \ourmethod a compelling solution to taskset generation in settings where additional inter-class information is hard or even impossible to obtain.
Conversely, we see that the easier tasksets generate by our method (the ones with 0.04 divergence) are significantly easier, sometimes even approaching purely random assignments. (\ie with a divergence of 0)

\subsection{Is a Good Embedding Really Enough?}


\begin{table}[h]
    \vspace{-1em}
    \centering
    \caption{
        \small
        Slope of the regression line between divergence and accuracy (in \% points) for different methods.
        MAML degrades at slower rate than metric-based methods, suggesting that it is better suited when transfer is challenging.
    }
    \label{tMethodSlope}
    \setlength{\tabcolsep}{3pt}
    \vspace{-1em}
    {\small
    \begin{tabular}{l cccc}
        \addlinespace
        \toprule


                        & CIFAR100   & mini-ImageNet & LFW10     & EMNIST \\
        \midrule
        MAML            & \h{-11.53} & \h{-6.68}     & \h{-3.10} & \h{-2.55} \\
        ANIL            & -12.44     & -7.32         & -3.61     & -5.70   \\
        ProtoNet        & -14.88     & -7.18         & -4.45     & -4.71   \\
        Multiclass      & -12.41     & -7.41         & -4.91     & -3.23   \\

        \bottomrule
    \end{tabular}
    }
    \vspace{-1em}
\end{table}

A closer inspection of the results in Table~\ref{tAccuracyComparison} hints at an unexpected trend: \emph{MAML becomes increasingly more competitive as the transfer from train to test classes become more challenging.}
To verify this hypothesis, we train all 4 few-shot algorithms on all tasksets of 4 datasets highlighted in Section~\ref{sControllingInformationOverlap}; then, we compute the slope of the regression line between divergence and test accuracies.
Table~\ref{tMethodSlope} reports those slopes, and confirms our hypothesis: gradient-based methods degrade slower than metric-based ones.

In turn, those experiments suggest a follow-up hypothesis: \emph{when train-test transfer is challenging enough, methods that adapt their embedding function should outperform those that do not}.
We verify this hypothesis in Table~\ref{tHardPartitions} where we compare accuracies of each algorithms on the most challenging taskset of each dataset.
We also include a \emph{Finetune} entry, which corresponds to Multiclass with the embedding function updated by 5 gradient steps on test tasks.
Confirming our hypothesis, MAML and Finetune dominate on all datasets.


\section{Conclusion}
\label{sDiscussion}
This manuscript focuses on the evaluation and analysis of few-shot classification methods.
Accordingly, we propose \ourmethod to generate tasksets of desired transfer difficulty, and with no requirement for additional human information about classes or their relationships.
After empiricaly validating \ourmethod, we generate tasksets to study the two main families of few-shot classification algorithms: gradient-based and metric-based.
As opposed to recent work suggesting that a good feature extractor might be enough for few-shot classification~\cite{Tian2020-gv, Raghu2019-ff}, we find that gradient-based methods outperform metric-based ones, especially when transfer is challenging.

Although seemingly contradicting, we believe the two hypotheses are compatible: when applied out of domain the metric implicitly learned by ProtoNet and Multiclass will likely require adaptation, \eg to adjust for domain shift or to discover new features.
For similar reasons and as highlighted in our experiments, ANIL's approximation of MAML will break since it does not adapt its feature extractor to avoid expensive second-order derivatives.

On the other hand, those methods don't have to pay the price of adaptation when knowledge transfer is sufficient to reach good performance, and can thus be scald to much larger datasets.
In fact, when test tasks are similar to training, MAML's adaptation by gradient descent can even lead to overfitting especially when working with limited labelled data.

We hope our contribution can help researchers analyse few-shot learning methods and answer some of the above questions.
Future research directions for \ourmethod include extensions to regression and multi-label classification tasks.

\subsubsection*{Acknowledgements}
Fei Sha is on leave from University of Southern California.
This work is partially supported by NSF AwardsIIS-1513966/ 1632803/1833137, CCF-1139148, DARPA Award\#: FA8750-18-2-0117, DARPA-D3M - Award UCB-00009528, Google Research Awards, gifts from Facebook and Netflix, and ARO\# W911NF-12-1-0241 and W911NF-15-1-0484.

\clearpage
\pagebreak

{\small
\bibliographystyle{ieee_fullname}
\bibliography{ms}
}

\clearpage
\pagebreak
\appendix

\section{Supplementary Material}

The following sections provide additional details on our experiments, our implementations, and the tasksets generated with our method.

\subsection{Experimental Details}

\subsection{Controlling Information Overlap}
For results reported in Figure~\ref{fMIAcc} we train the Multiclass network for 200 epochs on the train classes, and measure test accuracy over 600 test tasks with the weights obtained at the best validation epoch.
For optimization, we use Adam with a tuned learning rate and other hyper-parameters set to the default values in PyTorch.
We do not use weight decay regluarization nor decay the learning rate, and set the mini-batch size to 64.

For the PCA projections of Figure~\ref{fPCAClustering}, we use the PCA implementation in scikit-learn with default hyper-parameters and simply set \texttt{ncomponents=2}.
When reporting $\vert \vert \mutrain - \mutest \vert\vert$, we measure the Euclidean distance in the original embedding space not the projected one.

Since we do not need to optimize their parameters, we use Gaussians to measure correlation between divergence and accuracies in Table~\ref{tMeasureOverlap}.
We estimate the means and covariance from the class embeddings, and use a small diagonal damping factor of 0.001 to ensure numerical stability.

\subsection{Comparing Semantic vs Embedding Clusters}
For Table~\ref{tAccuracyComparison}, the label \emph{Easy} corresponds to $D = 0.04$, \emph{Hard} to $D = 0.96$.
For \emph{Random}, we randomly partition classes among train, validation, and test sets which is equivalent to $D = 0$.
The CNN4 and ResNet12 implementeations are provided with our code release.

When comparing clustering trees in Table~\ref{tHopDistances}, we use the agglomerative clustering implementation found in scikit-learn, and set the linkage argument to \texttt{'ward'}.
We parse the structures in the following 2 files to build the WordNet graph over ImageNet-1k:
\begin{itemize}
    \small
    \item \url{http://www.image-net.org/api/xml/ReleaseStatus.xml} and
    \item \url{http://www.image-net.org/api/xml/structure_released.xml},
\end{itemize}
and prune the graph to discard leaves irrelevant to ImageNet-1k synsets.

\subsection{Is a Good Embedding Really Enough?}
For each method, the slopes in Table~\ref{tMethodSlope} are computed using the test accuracies when varying $R$ from 0.04 to 0.96.
In Table~\ref{tHardPartitions}, we report the test accuracies when $D = 0.96$.

\subsection{Code Release}
With this submission, we include implementation for \ourmethod, as well as parallelized implementations for popular few-shot learning algorithms.
(MAML, ANIL, and ProtoNet)

The few-shot learning implementations proceed one task at a time.
Before processing the task, the model is replicated across all GPUs; then, the data of the task is split across GPUs and processed by each of the replicas independently.
(This is parallelization is made simple thanks to PyTorch's \texttt{DataParallel} wrapper.)
Finally, the output of each replica is gathered on the first GPU to compute the cross-entropy loss.
Those implementations are available in \texttt{maml.py}, \texttt{anil.py}, and \texttt{protonet.py}.

The \ourmethod implementation is provided in \texttt{atg.py}.
It (and our other implementations) can be ran as-is, using the python command:
\begin{verbatim}
    python cvpr_code/atg.py
\end{verbatim}
For more details on our implementations, please refer to the included \texttt{README.md} file.
As mentioned in the main text, our implementations are available at

\noindent
\url{\codelink}.

\subsubsection{Software Acknowledgement}
This manuscript is made possible thanks to the following open-source software: Python~\cite{python}, Numpy~\cite{numpy}, PyTorch~\cite{NEURIPS2019_9015}, torchvision~\cite{torchvision}, scikit-learn~\cite{scikit-learn}, learn2learn~\cite{Arnold2020-ss}, and matplotlib~\cite{Hunter:2007}.

\subsection{Taskset Partitions}
We describe the tasksets generated by our method for different degrees of difficulty, obtained for different values of $R = 0.04, 0.32, 0.64, 0.96$.
The following tables provide the train, validation, and test class partitions for
\begin{itemize}
    \item Table~\ref{tCifar100Partitions} for CIFAR100 as implemented in torchvision,
    \item Table~\ref{tMIPartitions} for mini-ImageNet as implemented in learn2learn,
    \item Table~\ref{tEmnistPartitions} for EMNIST as implemented in torchvision,
    \item Table~\ref{tLFW10Partitions} for LFW10 as implemented in scikit-learn, and
    \item Tables~\ref{tTIPartitions} and \ref{tTIPartitions2} for tiered-ImageNet as implemented in learn2learn.
\end{itemize}
All those partitions are obtained with \ourmethod, with the following implementation details.
First, we normalize the class embeddings to unit norm; then, optimize $\mutrain$ and $\mutest$ to minimize the objective $J$.
To that end, we use SGD with a $0.1$ learning rate and momentum set to $0.9$ for 7,000 iterations.


\begin{table*}[t]
    \vspace{0em}
    \centering
    {\small
    \centering
    \caption{
        \small
        Taskset partitions for CIFAR100.
    }
    \label{tCifar100Partitions}
    \vspace{-1em}
    \setlength{\tabcolsep}{3pt}
    \begin{tabular}{@{}ll p{0.85\linewidth} @{}}
        \toprule

        Setting    & Taskset    & Classes       \\
        \midrule

        D = 0.04   & Train      & 37, 39, 41, 89, 74, 15, 19, 8, 31, 95, 77, 38, 32, 17, 12, 90, 42, 84, 75, 25, 20, 0, 71, 69, 4, 86, 63, 80, 55, 40, 5, 50, 94, 87, 72, 34, 76, 13, 10, 81, 88, 58, 65, 22, 3, 66, 64, 61, 16, 28, 9, 2, 43, 21, 46, 97, 11, 35, 36, 98\\
                   & Validation & 47, 91, 56, 83, 59, 26, 1, 93, 27, 73, 14, 48, 30, 70, 68, 54, 67, 6, 85, 57\\
                   & Test       & 52, 82, 44, 96, 60, 33, 49, 45, 18, 7, 92, 29, 51, 78, 62, 79, 24, 23, 53, 99\\
        \midrule
        D = 0.32   & Train      & 99, 42, 53, 32, 23, 4, 75, 63, 48, 68, 80, 34, 50, 55, 85, 8, 95, 41, 72, 89, 0, 39, 37, 88, 84, 3, 17, 20, 66, 65, 43, 40, 90, 12, 25, 64, 69, 21, 71, 10, 28, 86, 97, 5, 61, 76, 94, 2, 35, 46, 11, 16, 22, 9, 87, 98, 13, 81, 36, 58\\
                   & Validation & 47, 44, 26, 7, 27, 18, 56, 14, 29, 1, 78, 62, 24, 60, 15, 49, 38, 67, 73, 57\\
                   & Test       & 52, 96, 33, 82, 59, 93, 91, 83, 45, 51, 79, 70, 92, 6, 54, 74, 19, 30, 77, 31\\
        \midrule
        D = 0.64   & Train      & 43, 32, 50, 55, 75, 3, 57, 30, 72, 60, 88, 53, 99, 21, 49, 66, 73, 23, 65, 8, 97, 64, 0, 35, 95, 41, 2, 85, 48, 68, 84, 89, 11, 98, 28, 46, 40, 39, 36, 20, 10, 37, 17, 61, 25, 69, 90, 71, 9, 16, 12, 86, 22, 5, 94, 76, 87, 81, 13, 58\\
                   & Validation & 96, 26, 33, 27, 93, 6, 78, 70, 14, 45, 62, 38, 83, 56, 92, 31, 77, 42, 67, 80\\
                   & Test       & 52, 47, 44, 7, 59, 51, 18, 29, 79, 24, 82, 15, 19, 74, 91, 1, 54, 4, 34, 63\\
        \midrule
        D = 0.96   & Train      & 55, 35, 92, 83, 72, 75, 82, 65, 1, 32, 56, 91, 2, 98, 64, 11, 67, 36, 46, 57, 30, 8, 53, 60, 23, 99, 49, 73, 41, 28, 89, 84, 0, 48, 85, 40, 95, 10, 39, 68, 20, 61, 25, 17, 37, 9, 16, 90, 86, 22, 69, 5, 94, 71, 12, 76, 58, 13, 81, 87\\
                   & Validation & 96, 19, 15, 31, 6, 7, 44, 51, 21, 27, 62, 24, 4, 93, 74, 54, 14, 42, 50, 63\\
                   & Test       & 52, 43, 38, 26, 47, 33, 34, 3, 70, 59, 78, 29, 79, 45, 80, 97, 88, 66, 18, 77\\

        \bottomrule
    \end{tabular}
    }
    \vspace{-1em}
\end{table*}


\begin{table*}[t]
    \vspace{0em}
    \centering
    {\small
    \centering
    \caption{
        \small
        Taskset partitions for mini-ImageNet.
    }
    \label{tMIPartitions}
    \vspace{-1em}
    \setlength{\tabcolsep}{3pt}
    \begin{tabular}{@{}ll p{0.85\linewidth} @{}}
        \toprule

        Setting    & Taskset    & Classes       \\
        \midrule

        D = 0.04   & Train      & 70, 87, 26, 31, 69, 78, 4, 73, 28, 83, 43, 46, 36, 14, 58, 49, 50, 8, 65, 9, 64, 2, 67, 39, 76, 88, 18, 91, 47, 42, 29, 68, 25, 12, 94, 99, 98, 11, 92, 61, 74, 71, 52, 77, 85, 37, 56, 66, 93, 5, 17, 97, 48, 24, 21, 20, 30, 95, 60, 84\\
                   & Validation & 35, 96, 53, 15, 63, 1, 23, 22, 54, 13, 82, 79, 40, 44, 51, 32, 6, 89, 55, 80\\
                   & Test       & 7, 33, 19, 86, 72, 27, 90, 59, 57, 45, 0, 41, 81, 34, 75, 38, 3, 10, 62, 16\\
        \midrule
        D = 0.32   & Train      & 40, 22, 76, 66, 75, 85, 2, 23, 58, 36, 44, 91, 48, 39, 94, 72, 97, 34, 74, 17, 53, 90, 25, 92, 24, 70, 51, 98, 84, 64, 60, 43, 37, 29, 89, 69, 52, 41, 47, 57, 73, 26, 96, 28, 93, 55, 46, 31, 32, 42, 38, 61, 49, 50, 77, 95, 71, 56, 30, 21\\
                   & Validation & 19, 15, 63, 10, 20, 86, 82, 5, 88, 1, 27, 33, 35, 45, 12, 11, 68, 79, 80, 4\\
                   & Test       & 3, 13, 62, 16, 59, 18, 7, 54, 14, 81, 8, 6, 9, 83, 99, 0, 87, 65, 67, 78\\
        \midrule
        D = 0.64   & Train      & 4, 66, 2, 80, 76, 60, 40, 48, 23, 91, 22, 97, 36, 94, 75, 44, 84, 25, 17, 39, 74, 24, 29, 92, 90, 98, 34, 64, 72, 37, 53, 51, 70, 43, 47, 52, 89, 69, 73, 26, 93, 41, 55, 42, 57, 28, 61, 38, 46, 96, 32, 49, 50, 31, 77, 95, 56, 71, 30, 21\\
                   & Validation & 19, 13, 63, 59, 16, 10, 54, 81, 14, 8, 7, 27, 11, 33, 67, 65, 35, 85, 78, 58\\
                   & Test       & 3, 62, 20, 15, 18, 86, 5, 82, 88, 1, 99, 12, 9, 83, 87, 68, 0, 6, 45, 79\\
        \midrule
        D = 0.96   & Train      & 4, 45, 25, 6, 76, 48, 66, 91, 23, 84, 29, 80, 97, 44, 40, 22, 24, 94, 92, 36, 74, 39, 90, 75, 17, 64, 51, 47, 37, 98, 89, 52, 53, 72, 42, 34, 43, 69, 70, 73, 55, 26, 38, 93, 57, 50, 61, 41, 49, 46, 95, 28, 96, 32, 77, 31, 30, 56, 71, 21\\
                   & Validation & 62, 20, 59, 86, 81, 18, 16, 82, 88, 8, 12, 27, 67, 87, 33, 83, 65, 0, 78, 58\\
                   & Test       & 3, 19, 63, 13, 15, 5, 10, 54, 1, 14, 99, 85, 9, 68, 79, 7, 60, 11, 2, 35\\

        \bottomrule
    \end{tabular}
    }
    \vspace{-1em}
\end{table*}


\begin{table*}[t]
    \vspace{0em}
    \centering
    {\small
    \centering
    \caption{
        \small
        Taskset partitions for EMNIST.
    }
    \label{tEmnistPartitions}
    \vspace{-1em}
    \setlength{\tabcolsep}{3pt}
    \begin{tabular}{@{}ll p{0.85\linewidth} @{}}
        \toprule

        Setting    & Taskset    & Classes       \\
        \midrule

        D = 0.04   & Train      & 61, 52, 17, 45, 8, 32, 13, 36, 44, 1, 19, 56, 60, 0, 18, 24, 9, 16, 11, 50, 22, 12, 57, 53, 38, 6, 33, 48, 27, 58, 41, 59, 49, 10, 54, 40, 28, 3\\
                   & Validation & 55, 20, 7, 46, 42, 23, 34, 30, 31, 26, 14, 15\\
                   & Test       & 29, 21, 5, 35, 39, 25, 51, 37, 4, 2, 43, 47\\
        \midrule
        D = 0.32   & Train      & 31, 9, 17, 27, 34, 40, 14, 11, 4, 55, 57, 46, 5, 2, 60, 10, 61, 21, 53, 16, 8, 15, 39, 43, 19, 37, 33, 59, 41, 45, 6, 47, 54, 44, 18, 1, 3, 28\\
                   & Validation & 48, 0, 24, 7, 25, 32, 56, 36, 42, 20, 38, 35\\
                   & Test       & 22, 29, 50, 51, 30, 23, 26, 13, 58, 49, 12, 52\\
        \midrule
        D = 0.64   & Train      & 23, 28, 35, 17, 19, 41, 30, 47, 61, 40, 1, 44, 56, 18, 49, 27, 32, 11, 8, 45, 43, 42, 9, 58, 37, 13, 3, 16, 52, 24, 2, 50, 0, 6, 10, 39, 36, 26\\
                   & Validation & 5, 20, 60, 53, 46, 55, 21, 33, 4, 48, 12, 51\\
                   & Test       & 29, 34, 31, 14, 7, 57, 15, 22, 25, 59, 54, 38\\
        \midrule
        D = 0.96   & Train      & 19, 37, 1, 61, 18, 44, 45, 8, 25, 16, 6, 30, 11, 17, 40, 43, 3, 22, 51, 48, 42, 56, 23, 9, 27, 52, 2, 13, 39, 49, 32, 24, 58, 50, 0, 10, 26, 36\\
                   & Validation & 29, 34, 54, 20, 15, 55, 53, 7, 33, 41, 38, 47\\
                   & Test       & 5, 21, 14, 28, 60, 31, 46, 57, 4, 12, 59, 35\\

        \bottomrule
    \end{tabular}
    }
    \vspace{-1em}
\end{table*}


\begin{table*}[t]
    \vspace{0em}
    \centering
    {\small
    \centering
    \caption{
        \small
        Taskset partitions for LFW10.
    }
    \label{tLFW10Partitions}
    \vspace{-1em}
    \setlength{\tabcolsep}{3pt}
    \begin{tabular}{@{}ll p{0.85\linewidth} @{}}
        \toprule

        Setting    & Taskset    & Classes       \\
        \midrule

        D = 0.04   & Train      & 128, 133, 13, 100, 7, 67, 91, 140, 122, 66, 22, 6, 5, 42, 88, 0, 156, 107, 81, 149, 41, 148, 2, 80, 153, 62, 98, 83, 71, 123, 109, 58, 53, 84, 25, 35, 136, 8, 141, 146, 57, 56, 135, 157, 129, 143, 34, 117, 93, 4, 52, 118, 137, 73, 126, 15, 101, 97, 26, 138, 60, 50, 147, 127, 95, 104, 78, 134, 119, 121, 132, 33, 139, 31, 54, 124, 96, 30, 87, 44, 11, 27, 103, 39, 14, 38, 142, 89, 74, 70, 125, 69, 152, 23, 116, 120\\
                   & Validation & 63, 36, 111, 47, 9, 61, 28, 40, 68, 45, 21, 51, 108, 10, 105, 77, 75, 115, 144, 106, 12, 59, 37, 86, 130, 113, 49, 114, 46, 112, 16\\
                   & Test       & 82, 65, 76, 92, 90, 48, 110, 20, 17, 72, 79, 24, 3, 94, 18, 150, 29, 19, 145, 99, 64, 55, 155, 131, 154, 1, 43, 32, 85, 151, 102\\
        \midrule
        D = 0.32   & Train      & 77, 109, 122, 42, 40, 94, 85, 64, 2, 128, 81, 80, 118, 55, 17, 97, 54, 107, 153, 123, 83, 30, 20, 71, 25, 135, 56, 9, 156, 58, 101, 137, 96, 34, 115, 0, 117, 31, 22, 78, 138, 57, 62, 126, 38, 52, 27, 146, 129, 11, 144, 35, 41, 91, 143, 124, 114, 33, 132, 60, 95, 50, 87, 6, 147, 127, 134, 44, 8, 84, 53, 73, 93, 103, 139, 98, 157, 121, 89, 26, 119, 14, 4, 141, 69, 15, 23, 104, 116, 142, 152, 125, 39, 120, 74, 70\\
                   & Validation & 48, 110, 92, 99, 82, 37, 111, 45, 59, 67, 1, 88, 19, 12, 13, 108, 140, 3, 24, 131, 133, 66, 79, 72, 90, 29, 130, 16, 7, 76, 150\\
                   & Test       & 155, 68, 36, 105, 102, 28, 63, 18, 51, 113, 47, 75, 145, 43, 5, 65, 61, 106, 100, 10, 86, 154, 148, 46, 112, 136, 21, 149, 32, 151, 49\\
        \midrule
        D = 0.64   & Train      & 81, 16, 49, 150, 66, 21, 137, 25, 52, 102, 71, 54, 22, 58, 1, 136, 97, 118, 107, 99, 83, 40, 111, 57, 17, 96, 33, 134, 56, 62, 117, 87, 123, 133, 115, 138, 156, 78, 38, 64, 34, 80, 60, 2, 42, 144, 20, 124, 76, 31, 35, 50, 147, 9, 6, 0, 89, 126, 44, 143, 103, 116, 26, 91, 127, 84, 95, 85, 93, 146, 23, 114, 139, 53, 73, 8, 98, 141, 14, 119, 41, 15, 121, 132, 142, 69, 125, 152, 4, 104, 39, 129, 157, 120, 74, 70\\
                   & Validation & 36, 59, 68, 75, 29, 61, 149, 51, 140, 67, 113, 110, 154, 153, 148, 122, 77, 79, 19, 135, 145, 24, 94, 65, 130, 32, 151, 27, 82, 46, 108\\
                   & Test       & 28, 155, 92, 13, 106, 131, 3, 18, 5, 7, 47, 100, 12, 63, 45, 112, 72, 30, 105, 43, 48, 55, 88, 128, 37, 101, 10, 109, 90, 86, 11\\
        \midrule
        D = 0.96   & Train      & 56, 83, 62, 43, 55, 110, 30, 87, 107, 57, 134, 16, 133, 64, 101, 42, 82, 66, 99, 137, 34, 20, 50, 111, 25, 96, 156, 2, 79, 76, 117, 11, 91, 33, 0, 32, 27, 10, 6, 53, 138, 102, 144, 26, 71, 31, 9, 116, 54, 98, 60, 97, 40, 114, 39, 15, 136, 143, 8, 89, 141, 123, 35, 80, 103, 84, 142, 38, 147, 78, 119, 95, 124, 93, 157, 74, 104, 41, 139, 146, 126, 73, 121, 127, 152, 4, 85, 23, 44, 132, 125, 129, 120, 14, 69, 70\\
                   & Validation & 36, 92, 28, 106, 72, 140, 12, 24, 18, 100, 3, 75, 61, 63, 5, 47, 48, 17, 45, 88, 49, 90, 1, 22, 37, 19, 148, 58, 115, 130, 109\\
                   & Test       & 7, 155, 59, 67, 94, 113, 149, 13, 131, 105, 135, 65, 122, 29, 68, 51, 21, 108, 150, 153, 151, 145, 128, 154, 52, 112, 86, 77, 81, 46, 118\\

        \bottomrule
    \end{tabular}
    }
    \vspace{-1em}
\end{table*}


\begin{table*}[t]
    \vspace{0em}
    \centering
    {\small
    \centering
    \caption{
        \small
        Taskset partitions for tiered-ImageNet.
    }
    \label{tTIPartitions}
    \vspace{-1em}
    \setlength{\tabcolsep}{3pt}
    \begin{tabular}{@{}ll p{0.85\linewidth} @{}}
        \toprule

        Setting    & Taskset    & Classes       \\
        \midrule

        D = 0.04   & Train      & 163, 403, 375, 148, 381, 431, 160, 363, 187, 499, 412, 321, 103, 264, 553, 318, 560, 410, 448, 97, 173, 179, 368, 331, 442, 557, 165, 457, 376, 337, 116, 325, 91, 598, 227, 559, 150, 77, 36, 118, 104, 411, 212, 8, 9, 188, 418, 278, 509, 498, 383, 82, 508, 473, 468, 266, 133, 310, 138, 436, 385, 423, 584, 490, 263, 520, 358, 238, 195, 297, 510, 193, 497, 141, 354, 283, 391, 426, 320, 299, 393, 32, 356, 603, 433, 5, 400, 192, 29, 251, 486, 512, 382, 384, 157, 558, 137, 555, 366, 482, 117, 265, 170, 67, 446, 105, 447, 319, 352, 462, 351, 42, 336, 268, 541, 52, 394, 539, 593, 342, 456, 216, 151, 214, 416, 276, 296, 361, 395, 98, 581, 196, 369, 249, 554, 390, 537, 570, 396, 267, 569, 261, 576, 204, 181, 286, 115, 220, 126, 480, 347, 16, 30, 346, 322, 524, 145, 538, 277, 372, 22, 566, 516, 84, 246, 229, 496, 59, 589, 472, 112, 125, 406, 159, 288, 518, 421, 466, 7, 475, 38, 211, 111, 69, 350, 364, 444, 262, 443, 129, 333, 20, 328, 329, 134, 463, 123, 89, 85, 275, 341, 491, 100, 314, 389, 450, 437, 607, 335, 224, 345, 451, 413, 428, 551, 128, 174, 359, 483, 71, 289, 580, 386, 234, 404, 340, 75, 536, 543, 591, 365, 408, 217, 106, 18, 215, 155, 15, 92, 198, 432, 392, 454, 306, 79, 168, 176, 237, 21, 343, 371, 47, 152, 547, 525, 94, 316, 33, 63, 334, 99, 527, 360, 53, 402, 323, 419, 272, 37, 597, 164, 529, 605, 434, 449, 532, 517, 295, 500, 202, 80, 110, 130, 144, 357, 573, 201, 146, 344, 96, 315, 332, 552, 549, 60, 250, 535, 232, 222, 167, 226, 519, 544, 355, 308, 252, 131, 269, 158, 46, 515, 414, 76, 61, 312, 28, 132, 240, 182, 301, 501, 489, 505, 338, 567, 210, 439, 142, 467, 190, 50, 578, 180, 565, 487, 455, 546, 19, 253, 40, 285, 41, 317, 175, 293, 225, 161, 495, 239, 219, 424, 25, 184, 279, 247, 542, 572, 207, 31, 587, 66, 54, 185, 550, 453, 563\\
                   & Validation & 588, 55, 504, 493, 209, 136, 460, 399, 2, 113, 339, 292, 503, 484, 459, 122, 90, 304, 156, 313, 245, 415, 135, 166, 602, 464, 4, 213, 307, 291, 430, 401, 441, 548, 526, 58, 273, 465, 282, 13, 330, 425, 154, 49, 458, 178, 68, 601, 119, 388, 203, 86, 230, 513, 294, 81, 280, 305, 481, 88, 309, 420, 51, 574, 139, 189, 507, 17, 143, 87, 586, 438, 200, 14, 248, 387, 577, 440, 257, 600, 521, 422, 169, 534, 78, 575, 397, 107, 44, 10, 101, 471, 523, 327, 405, 108, 72, 56, 595, 579, 311, 378, 26, 6, 284, 485, 43, 377, 303, 221, 242, 102, 429, 208, 147, 290, 177, 479, 109, 27, 120\\
                   & Test       & 445, 583, 348, 259, 199, 93, 162, 514, 604, 571, 461, 469, 300, 585, 545, 74, 530, 23, 62, 228, 582, 511, 494, 83, 270, 271, 353, 470, 256, 594, 435, 281, 606, 274, 3, 191, 153, 124, 528, 452, 502, 260, 476, 374, 409, 171, 599, 540, 236, 298, 57, 326, 596, 258, 140, 255, 35, 287, 1, 205, 362, 522, 194, 45, 114, 370, 561, 218, 183, 233, 206, 65, 568, 186, 417, 592, 492, 24, 243, 127, 427, 95, 64, 380, 302, 254, 478, 590, 73, 367, 379, 121, 556, 149, 488, 11, 477, 172, 223, 197, 39, 407, 373, 231, 474, 48, 235, 244, 506, 349, 562, 0, 324, 241, 564, 34, 533, 70, 12, 531, 398\\
        \midrule
        D = 0.32   & Train      & 256, 34, 135, 249, 17, 322, 229, 403, 511, 234, 422, 464, 468, 215, 287, 198, 16, 18, 122, 452, 482, 84, 423, 12, 265, 481, 439, 59, 27, 426, 362, 33, 416, 208, 20, 512, 280, 361, 513, 237, 446, 288, 314, 411, 559, 421, 574, 92, 309, 52, 86, 383, 522, 168, 340, 320, 507, 306, 323, 365, 497, 103, 569, 186, 582, 181, 604, 410, 506, 562, 330, 71, 364, 354, 193, 579, 264, 75, 457, 195, 85, 390, 42, 328, 341, 176, 461, 286, 396, 311, 310, 98, 112, 272, 67, 523, 8, 530, 470, 69, 370, 592, 120, 108, 499, 393, 350, 127, 601, 76, 137, 428, 73, 160, 212, 442, 232, 22, 57, 472, 585, 316, 543, 595, 220, 602, 520, 124, 68, 105, 369, 503, 502, 87, 534, 177, 325, 389, 599, 381, 179, 531, 458, 427, 116, 515, 21, 134, 456, 4, 5, 91, 333, 117, 319, 404, 261, 145, 576, 568, 53, 560, 133, 366, 571, 477, 216, 372, 387, 269, 167, 490, 524, 516, 344, 268, 15, 30, 106, 510, 538, 408, 437, 359, 494, 541, 539, 412, 544, 152, 444, 118, 475, 164, 414, 80, 447, 518, 329, 243, 241, 151, 210, 600, 533, 238, 360, 297, 591, 432, 31, 41, 558, 448, 252, 386, 315, 148, 462, 121, 352, 555, 556, 607, 590, 371, 476, 551, 201, 305, 170, 554, 240, 225, 224, 37, 537, 596, 174, 606, 454, 278, 449, 47, 501, 480, 50, 345, 150, 392, 97, 338, 196, 79, 581, 163, 89, 406, 184, 146, 566, 451, 593, 96, 312, 580, 219, 129, 474, 155, 498, 394, 38, 295, 419, 110, 28, 536, 573, 400, 443, 402, 491, 605, 308, 525, 463, 19, 128, 109, 517, 483, 54, 123, 466, 335, 587, 301, 40, 552, 334, 343, 285, 336, 46, 519, 547, 130, 489, 589, 161, 202, 158, 332, 222, 565, 131, 250, 94, 277, 226, 450, 357, 487, 424, 527, 535, 60, 144, 500, 247, 185, 467, 597, 132, 567, 532, 182, 578, 61, 142, 279, 505, 289, 529, 455, 355, 63, 25, 549, 546, 550, 180, 563, 317, 190, 239, 293, 253, 542, 495, 572, 66, 453, 175, 207\\
                   & Validation & 35, 119, 55, 88, 540, 300, 6, 267, 93, 259, 156, 379, 588, 284, 39, 307, 445, 430, 0, 492, 431, 62, 235, 397, 299, 283, 349, 23, 221, 147, 281, 140, 83, 233, 10, 251, 488, 526, 318, 391, 90, 7, 303, 100, 263, 188, 376, 45, 356, 178, 586, 509, 290, 9, 435, 324, 471, 291, 388, 154, 139, 32, 436, 385, 374, 548, 260, 496, 504, 77, 363, 417, 377, 203, 282, 577, 557, 337, 564, 294, 373, 81, 29, 200, 58, 545, 78, 399, 514, 2, 102, 384, 24, 171, 327, 484, 321, 358, 465, 351, 169, 153, 56, 347, 275, 14, 246, 469, 197, 214, 254, 440, 257, 434, 266, 95, 583, 211, 204, 101, 138\\
                   & Test       & 460, 401, 348, 149, 292, 228, 244, 166, 107, 236, 3, 273, 143, 528, 43, 258, 274, 51, 162, 70, 230, 313, 113, 485, 104, 74, 209, 242, 304, 353, 271, 245, 111, 270, 378, 441, 114, 205, 136, 36, 493, 479, 433, 13, 331, 409, 213, 594, 183, 429, 65, 126, 49, 64, 302, 425, 172, 187, 173, 508, 99, 382, 575, 262, 478, 82, 115, 486, 255, 407, 189, 248, 217, 346, 199, 276, 191, 598, 194, 584, 438, 231, 48, 603, 553, 72, 521, 459, 206, 11, 44, 368, 223, 415, 296, 165, 405, 418, 326, 570, 298, 1, 125, 218, 342, 192, 141, 413, 398, 420, 375, 157, 159, 227, 473, 339, 26, 380, 395, 561, 367\\

        \bottomrule
    \end{tabular}
    }
    \vspace{-1em}
\end{table*}

\begin{table*}[t]
    \vspace{0em}
    \centering
    {\small
    \centering
    \caption{
        \small
        Taskset partitions for tiered-ImageNet. (continued)
    }
    \label{tTIPartitions2}
    \vspace{-1em}
    \setlength{\tabcolsep}{3pt}
    \begin{tabular}{@{}ll p{0.85\linewidth} @{}}
        \toprule

        Setting    & Taskset    & Classes       \\
        \midrule

        D = 0.64   & Train      & 14, 582, 59, 559, 181, 209, 323, 506, 464, 562, 76, 524, 310, 316, 352, 65, 254, 216, 29, 516, 89, 298, 520, 177, 235, 101, 360, 274, 427, 344, 283, 230, 138, 302, 509, 100, 147, 246, 231, 227, 457, 217, 229, 187, 37, 232, 173, 264, 127, 27, 99, 322, 238, 601, 371, 168, 472, 103, 437, 287, 225, 160, 53, 164, 574, 473, 518, 479, 73, 604, 52, 591, 77, 599, 170, 359, 135, 263, 115, 513, 241, 454, 327, 501, 275, 330, 502, 576, 350, 98, 272, 538, 406, 558, 17, 198, 42, 475, 195, 432, 108, 21, 30, 153, 426, 151, 311, 339, 47, 499, 22, 114, 534, 449, 67, 333, 592, 105, 548, 560, 593, 523, 120, 556, 533, 461, 436, 381, 262, 86, 122, 428, 96, 503, 462, 353, 197, 402, 392, 607, 374, 291, 544, 57, 458, 444, 551, 296, 331, 117, 510, 102, 470, 541, 602, 145, 186, 585, 466, 320, 571, 118, 554, 606, 471, 555, 490, 134, 5, 208, 566, 319, 443, 226, 373, 537, 587, 387, 306, 600, 477, 308, 212, 31, 404, 564, 596, 326, 176, 44, 269, 224, 312, 605, 476, 38, 439, 400, 494, 278, 91, 268, 146, 552, 480, 590, 578, 196, 345, 286, 4, 125, 92, 261, 434, 41, 192, 28, 565, 159, 40, 243, 112, 97, 211, 50, 440, 71, 386, 280, 412, 517, 33, 569, 498, 500, 332, 366, 581, 358, 110, 394, 121, 19, 116, 297, 519, 511, 309, 8, 338, 474, 129, 169, 295, 408, 79, 487, 357, 343, 536, 185, 219, 403, 305, 354, 68, 491, 535, 579, 512, 152, 505, 525, 252, 315, 184, 109, 563, 543, 515, 240, 361, 463, 595, 546, 550, 130, 522, 456, 128, 539, 54, 396, 372, 334, 179, 106, 568, 148, 336, 15, 450, 567, 329, 547, 335, 419, 202, 132, 589, 123, 46, 448, 542, 451, 201, 150, 182, 370, 549, 131, 161, 142, 155, 279, 580, 317, 63, 174, 163, 222, 285, 66, 250, 424, 190, 572, 61, 453, 94, 239, 277, 573, 60, 489, 25, 293, 355, 253, 289, 495, 301, 158, 455, 247, 180, 483, 467, 144, 527, 597, 532, 175, 529, 207\\
                   & Validation & 348, 156, 3, 378, 259, 356, 379, 162, 313, 13, 588, 74, 236, 324, 304, 55, 166, 303, 445, 35, 460, 9, 214, 508, 486, 405, 104, 242, 351, 478, 276, 376, 149, 488, 496, 594, 409, 459, 398, 603, 465, 140, 526, 20, 281, 299, 39, 48, 136, 492, 215, 570, 255, 233, 493, 540, 341, 204, 365, 364, 414, 88, 325, 62, 411, 270, 416, 415, 388, 75, 257, 83, 346, 69, 133, 16, 249, 413, 248, 199, 137, 194, 507, 294, 7, 258, 553, 188, 2, 531, 410, 271, 1, 342, 446, 203, 256, 221, 431, 126, 172, 34, 363, 307, 395, 265, 468, 218, 328, 273, 24, 80, 390, 481, 266, 442, 87, 139, 530, 447, 141\\
                   & Test       & 228, 6, 401, 300, 51, 93, 429, 23, 385, 349, 143, 441, 245, 64, 70, 586, 377, 78, 213, 545, 205, 528, 430, 72, 45, 282, 18, 485, 11, 82, 384, 81, 56, 84, 178, 119, 113, 314, 575, 423, 418, 375, 368, 407, 292, 267, 340, 598, 90, 49, 417, 584, 399, 244, 438, 191, 165, 157, 223, 284, 124, 43, 433, 183, 557, 521, 32, 154, 504, 260, 318, 382, 107, 362, 234, 383, 367, 422, 435, 171, 12, 210, 10, 189, 111, 452, 237, 420, 290, 380, 514, 484, 193, 36, 425, 206, 220, 85, 561, 421, 347, 167, 393, 58, 497, 0, 583, 251, 577, 26, 200, 389, 469, 482, 397, 288, 95, 321, 369, 391, 337\\
        \midrule
        D = 0.96   & Train      & 600, 509, 217, 138, 225, 220, 481, 164, 341, 565, 69, 359, 135, 602, 430, 409, 193, 550, 435, 103, 380, 498, 104, 390, 265, 479, 215, 392, 17, 488, 375, 268, 368, 535, 505, 461, 99, 101, 198, 254, 147, 287, 476, 546, 494, 236, 556, 592, 276, 100, 145, 281, 31, 339, 153, 487, 146, 14, 519, 58, 496, 477, 563, 230, 517, 9, 433, 115, 203, 28, 24, 492, 562, 555, 141, 425, 311, 274, 523, 57, 534, 591, 302, 369, 574, 291, 485, 21, 235, 181, 208, 478, 186, 80, 395, 273, 475, 130, 22, 117, 283, 333, 264, 32, 490, 263, 538, 551, 120, 132, 548, 458, 110, 436, 212, 411, 266, 312, 590, 559, 352, 197, 374, 278, 102, 231, 486, 321, 189, 581, 322, 387, 406, 50, 363, 389, 553, 542, 41, 344, 575, 508, 308, 447, 91, 347, 330, 114, 306, 182, 373, 5, 66, 499, 59, 280, 501, 474, 288, 323, 1, 560, 463, 86, 172, 350, 338, 176, 320, 439, 142, 317, 468, 112, 450, 105, 319, 427, 393, 549, 262, 432, 19, 27, 598, 40, 491, 71, 331, 256, 129, 402, 453, 440, 168, 219, 584, 564, 10, 53, 159, 525, 8, 237, 44, 537, 326, 134, 92, 196, 589, 242, 122, 353, 547, 296, 68, 342, 394, 567, 209, 202, 603, 309, 434, 211, 480, 426, 536, 345, 577, 337, 299, 570, 96, 109, 121, 272, 444, 190, 47, 471, 305, 240, 192, 261, 297, 185, 428, 29, 125, 200, 569, 403, 184, 286, 152, 42, 572, 243, 127, 354, 252, 54, 544, 221, 33, 381, 148, 511, 79, 334, 400, 539, 497, 327, 131, 106, 229, 150, 456, 269, 279, 579, 358, 372, 336, 595, 451, 412, 232, 443, 543, 522, 179, 473, 224, 295, 52, 163, 361, 169, 512, 123, 201, 128, 568, 15, 515, 404, 396, 239, 580, 366, 4, 357, 495, 329, 472, 187, 173, 335, 293, 222, 448, 253, 386, 46, 77, 174, 155, 573, 370, 275, 61, 25, 161, 408, 158, 97, 285, 489, 116, 315, 277, 250, 467, 94, 355, 301, 455, 247, 424, 60, 180, 289, 144, 419, 483, 527, 175, 63, 597, 532, 529, 207\\
                   & Validation & 156, 259, 124, 23, 133, 20, 162, 89, 13, 245, 136, 356, 199, 45, 526, 531, 378, 191, 165, 514, 594, 82, 429, 84, 223, 282, 55, 95, 226, 314, 459, 248, 524, 241, 154, 318, 414, 34, 157, 260, 518, 171, 2, 26, 601, 292, 39, 582, 558, 90, 6, 36, 452, 464, 328, 449, 457, 85, 170, 284, 398, 37, 139, 216, 521, 12, 418, 56, 604, 111, 503, 38, 271, 73, 533, 376, 183, 270, 249, 587, 251, 137, 599, 607, 118, 462, 316, 332, 346, 540, 437, 151, 107, 500, 246, 143, 420, 557, 585, 160, 605, 484, 520, 98, 552, 294, 206, 576, 30, 233, 383, 438, 367, 513, 482, 258, 442, 388, 195, 343, 566\\
                   & Test       & 348, 228, 3, 93, 325, 545, 78, 213, 324, 465, 300, 303, 64, 11, 166, 140, 81, 51, 205, 588, 493, 401, 18, 460, 238, 214, 530, 35, 445, 177, 74, 583, 586, 385, 72, 507, 504, 48, 83, 469, 304, 516, 149, 188, 340, 113, 257, 119, 210, 49, 167, 466, 593, 415, 421, 365, 267, 362, 290, 416, 405, 204, 76, 7, 578, 441, 234, 255, 310, 218, 351, 87, 194, 16, 606, 423, 454, 244, 62, 379, 596, 399, 377, 384, 422, 88, 227, 502, 571, 391, 70, 431, 126, 554, 407, 528, 307, 413, 541, 364, 446, 43, 397, 108, 178, 0, 506, 75, 298, 67, 410, 360, 561, 65, 371, 510, 313, 417, 470, 349, 382\\

        \bottomrule
    \end{tabular}
    }
    \vspace{-1em}
\end{table*}

\end{document}